\documentclass{ws-ijait}

\usepackage[T1]{fontenc}
\usepackage{graphicx}
\usepackage{amsmath, amssymb}
\usepackage{url}
\usepackage{caption}
\usepackage{subcaption}
\usepackage{color}
\usepackage{array}
\usepackage{booktabs} 
\usepackage{multirow}
\usepackage{algorithm}
\usepackage[noend]{algpseudocode} 
\usepackage{colortbl}
\usepackage{amsfonts}
\usepackage{tikz}
\usetikzlibrary{shapes, arrows.meta, positioning, fit} 
\usepackage{pifont}
\usepackage[super]{cite}
\usepackage[inline]{enumitem} 
\usepackage[utf8]{inputenc}

\algnewcommand\And{\textbf{and}} 
\algnewcommand\Or{\textbf{or}}
\newcommand{\CL}{\mathcal{C}_L}

\newcommand{\CT}{\mathcal{C}_T}

\newcommand{\CLfinal}{\mathcal{C}_{L,final}}

\begin{document}

\markboth{Balafas, Tsouros, Ploskas, and Stergiou}
{Overcoming Over-Fitting in Global CA via Query-Driven Interactive
Refinement}

\catchline{}{}{}{}{}

\title{Overcoming Over-Fitting in Constraint Acquisition via Query-Driven Interactive Refinement}

\author{Vasileios Balafas}

\address{Department of Electrical and Computer Engineering, University of Western Macedonia\\
Campus ZEP, Kozani, 50100, Greece\\
v.balafas@uowm.gr}

\author{Dimos Tsouros}

\address{Department of Computer Science, KU Leuven\\
Celestijnenlaan 200a, Leuven, 3001, Belgium\\
dimos.tsouros@kuleuven.be}

\author{Nikolaos Ploskas}

\address{Department of Electrical and Computer Engineering, University of Western Macedonia\\
Campus ZEP, Kozani, 50100, Greece\\
nploskas@uowm.gr}

\author{Kostas Stergiou}

\address{Department of Electrical and Computer Engineering, University of Western Macedonia\\ Campus ZEP, Kozani, 50100, Greece\\
kstergiou@uowm.gr}

\maketitle

\begin{history}
\received{(Day Month Year)}
\revised{(Day Month Year)}
\accepted{(Day Month Year)}
\end{history}

\begin{abstract}
Manual modeling in Constraint Programming is a substantial bottleneck, which Constraint Acquisition (CA) aims to automate. However, passive CA methods are prone to over-fitting, often learning models that include spurious global constraints when trained on limited data, while purely active methods can be query-intensive. We introduce a hybrid CA framework specifically designed to address the challenge of over-fitting in CA. Our approach integrates passive learning for initial candidate generation, a query-driven interactive refinement phase that utilizes probabilistic confidence scores (initialized by machine learning priors) to systematically identify over-fitted constraints, and a specialized subset exploration mechanism to recover valid substructures from rejected candidates. A final active learning phase ensures model completeness. Extensive experiments on diverse benchmarks demonstrate that our interactive refinement phase is crucial for achieving high target model coverage and overall model accuracy from limited examples, doing so with manageable query complexity. This framework represents a substantial advancement towards robust and practical constraint acquisition in data-limited scenarios.

\end{abstract}

\section{Introduction}
\label{sec:intro}

Constraint Programming (CP) is a powerful paradigm for solving combinatorial problems across diverse applications~\cite{rossi2006handbook,baptiste2001constraint,lin2013csp}. However, its adoption is often hindered by the \emph{modeling bottleneck}~\cite{freuder2014grand,bessiere2016new,osullivan2010automated}: translating real-world requirements into a correct and efficient CP model requires significant expertise and time. This challenge is especially acute when dealing with expressive \emph{global constraints} such as \texttt{AllDifferent} or \texttt{Sum}~\cite{VanHoeve2006}, where an incorrect formulation can cripple a model.

Constraint Acquisition (CA)~\cite{Bessiere2017,DeRaedt_Passerini_Teso_2018} aims to automate the modeling process by learning the target constraint network from examples or user interaction. CA approaches range from \textbf{passive learning}~\cite{bessiere2005sat,BelSim}, which infers constraints from a fixed dataset but is prone to \textbf{over-fitting} due to the typically small number of examples~\cite{balafas2024impact}, to \textbf{active learning}, which queries an oracle to resolve ambiguities but can suffer from a high \textbf{query burden}~\cite{bessiere2013constraint,tsouros2018efficient}. \textbf{Hybrid methods}~\cite{balafashybrid} combine these, but they also lack mechanisms to specifically address the over-fitting of (global) constraints identified in the initial passive phase.

Over-fitting is a critical challenge that arises when passive learning identifies constraints that are consistent with the few available examples but are overly specific, capturing coincidental patterns rather than true problem rules. Accepting such constraints as part of the model, or passing them to an active learning phase, in the context of hybrid CA, can lead to bad models or inefficient refinement, while discarding them risks losing valuable structural information. This study introduces a framework to systematically identify and refine such over-fitted global constraints, leveraging the speed of initial pattern detection while handling the resulting inaccuracies through targeted oracle interaction.

Our research addresses the following questions:
\begin{enumerate}
    \item \textbf{RQ1:} How can potentially over-fitting sets of global constraints  be systematically evaluated and refined to converge towards the true constraints, preserving valid structures while eliminating spurious generalizations?
    \item \textbf{RQ2:} Which interactive strategies for query generation, confidence management, and resource allocation can most efficiently guide this refinement process, minimizing the interaction cost for the oracle?
\end{enumerate}

To address these questions, we have developed an enhanced hybrid CA framework. The process begins with (1) a passive learning phase that generates initial candidate (global) constraints from a few positive examples. The core of our method is (2) a Query-Driven Interactive Refinement stage. This phase vets each candidate constraint by assigning it a machine learning-based confidence score, and then posts targeted queries to an oracle. Based on the feedback, if a candidate constraint is rejected as over-fitted, a \emph{Subset Exploration} mechanism attempts to recover \emph{valid sub-constraints} (i.e., similar constraints on smaller sets of variables).
Finally, (3) the active learning phase learns any remaining simple constraints to complete the model.

Our primary contributions, which enable this CA process, include the following:
\begin{enumerate}
\item A three-phase hybrid architecture featuring a central Query-Driven Interactive Refinement phase that explicitly targets the rejection or refinement of over-fitted global constraints.
\item A specialized Subset Exploration mechanism that recovers valid substructures from rejected global constraints by modifying their scope while preserving parameters, thereby preventing information loss.
\item The integration of machine learning-based prior probability estimation and a Bayesian confidence framework to provide information about the probable existence of the learned constraints in the model, and principled evidence-based belief updates during interactive refinement.
\end{enumerate}
Empirical validation on diverse benchmarks demonstrated that our framework achieves high accuracy and target model coverage from limited examples with manageable query complexity, highlighting the crucial role of the interactive refinement phase in overcoming over-fitting. The remainder of this paper details our methodology (Section~\ref{sec:methodology}) after reviewing background and related work (Section~\ref{sec:background}-\ref{sec:relatedwork}), presents our experimental evaluation (Section~\ref{sec:experimental_setup}-\ref{sec:results}), and concludes with future directions (Section~\ref{sec:conclusion}).

\section{Background}
\label{sec:background}

\subsection{Constraint Satisfaction Problems (CSPs)}
A Constraint Satisfaction Problem (CSP)~\cite{rossi2006handbook} is formally defined as a triple $\mathcal{P} = (X, D, \mathcal{C})$, where $X = \{x_1, \ldots, x_n\}$ is a set of variables; $D = \{D(x_1), \ldots, D(x_n)\}$ is a set of corresponding domains of possible values; and $\mathcal{C} = \{c_1, \ldots, c_m\}$ is a set of constraints. Each constraint $c_j \in \mathcal{C}$ is a pair $(\text{vars}(c_j), \text{rel}(c_j))$, where its \emph{scope} $\text{vars}(c_j) \subseteq X$ is the set of variables it applies to, and its \emph{relation} $\text{rel}(c_j)$ specifies the allowed combinations of values for those variables. The number of variables in the scope, $|\text{vars}(c_j)|$, represents the constraint's \emph{arity}.

An \emph{assignment} maps variables to values from their domains; it is \emph{complete} if all variables in $X$ are assigned, and \emph{partial} otherwise. An assignment \emph{satisfies} a constraint if the combination of values for the variables in its scope is allowed by its relationship. A \emph{solution} to CSP is a complete assignment that satisfies all constraints in $\mathcal{C}$. The set of all solutions is denoted as $\text{sol}(\mathcal{C})$.

\subsection{Global Constraints}
\textit{Global constraints} represent complex combinatorial substructures involving an arbitrary number of variables (as opposed to \textit{fixed-arity} constraints such as binary ones) and are crucial for modeling expressiveness and solver efficiency~\cite{VanHoeve2006,beldiceanu2007global}. Our work focuses on acquiring models that include such constraints with the following examples:
\begin{itemize}
    \item \textbf{\texttt{AllDifferent}$(S)$}: Requires all variables in set $S$ to take distinct values.
    \item \textbf{\texttt{Sum}$(S, \bowtie, b)$}: Requires the sum of values for variables in $S$ to satisfy a relation $\bowtie \in \{=, \neq, <, \le, >, \ge\}$ with a bound $b$.
    \item \textbf{\texttt{Count}$(S, v, \bowtie, k)$}: Requires the number of variables in $S$ assigned value $v$ to satisfy a relation $\bowtie$ with a count $k$.
\end{itemize}

\subsection{Constraint Acquisition}
Constraint Acquisition (CA) aims to automatically learn a target constraint network $\CT$ given a vocabulary $(X, D)$, a language of possible constraint types $\Gamma$, and a source of information~\cite{bessiere2013constraint}. The set of all possible constraints that can be formed from $\Gamma$ over $(X,D)$ is  \emph{constraint bias} $B$. The information source can be a set of positive examples (solutions) $E^+$ and negative examples (non-solutions) $E^-$, or an \emph{oracle} $O$ that can classify assignments.

The goal is to learn a network $\CL \subseteq B$ such that its solution set is equivalent to the target's, i.e., $\text{sol}(\CL) = \text{sol}(\CT)$. It is assumed that $\CT$ is representable within the bias $B$. Interactive CA systems often maintain a probabilistic confidence score $P(c)$ for each candidate constraint $c \in B$, representing the belief that $c \in \CT$~\cite{tsouros2024learning}. This score is updated as new evidence is gathered from the oracle.

The interactive CA relies on Oracle $O$ to answer queries. The most common is the \emph{membership query} $\text{Ask}(O, Y)$, where the system presents an assignment $Y$ and the oracle returns 'Valid' if $Y \in \text{sol}(\CT)$ and 'Invalid' otherwise.

A crucial component is the \emph{query generation} strategy that aims to find informative assignments to pose to the oracle. Following established principles~\cite{bessiere2013constraint,tsouros2023guided}, an \textit{informative} query $Y$ satisfies all currently learned constraints but violates at least one remaining candidate constraint in the bias. It is typically formulated as an auxiliary CSP. To manage the complexity of this task, modern methods  such as \textbf{Projection-based Query Generation (PQ-GEN)}~\cite{tsouros2023guided} generate queries only over a subset of variables relevant to the current bias, simplifying the auxiliary CSP and improving efficiency. Our work adapts these principles to targeted query generation.

\section{Related Work}
\label{sec:relatedwork}

CA sits at the intersection of machine learning, constraint programming, and automated modeling \cite{DeRaedt_Passerini_Teso_2018}. The field aims to overcome modeling bottleneck \cite{osullivan2010automated} by learning constraints rather than requiring manual specification. Research has evolved along the passive, active, and hybrid lines.

\paragraph{Passive CA.} Learn from a fixed set of examples. ConAcq \cite{bessiere2005sat} employed a SAT-based version space algorithm, eliminating candidate constraints inconsistent with positive examples ($E^+$) and adding constraints required by negative examples ($E^-$). ModelSeeker \cite{BelSim} introduced pattern-based learning for global constraints, searching for structures such as `AllDifferent` or `Sum` patterns within $E^+$, often being effective with few examples but limited by its pattern catalogue. Recent studies have included COUNT-CP \cite{kumar2022learning} for learning numerical expressions, CLASSACQ \cite{prestwich2021classifier} using classifiers to score constraints, MINEACQ \cite{9643314} handling unlabeled data, and statistical approaches such as \cite{prestwich2024statistical}. The primary weakness remains over-fitting when the data are limited or non-representative \cite{balafas2024impact}.

\paragraph{Active CA.} Learn through interaction with an oracle using queries \cite{angluin1988queries}. ConAcq.2 \cite{Bessiere2017} added membership queries to the ConAcq. QuAcq \cite{bessiere2013constraint} improves efficiency by introducing partial queries and the FindScope procedure to identify violated constraints upon negative feedback. Extensions such as MultiAcq \cite{Arcangioli2016} and MQuAcq/MQuAcq-2 \cite{tsouros2018efficient,tsourosstructure} aimed to learn multiple constraints per negative query. Query generation strategies have been refined using projection \cite{tsouros2023guided} and probabilistic guidance \cite{tsouros2023guided,tsouros2024learning} (e.g., in GROWACQ),. The active learning literature \cite{Settles2009} offers relevant strategies such as uncertainty sampling or query-by-committee \cite{Seung1992,Freund1997}, though direct application to structured constraints can be complex. Furthermore, purely active approaches can struggle with learning global constraints from scratch becasue of the potentially exponential size of the search space (bias) required to represent all possible global constraints with their varying scopes and parameters, which, in turn, can lead to an intractable number of queries if such constraints are not pre-identified or if a more structured approach is not taken. Consequently, although accurate, active methods often face scalability issues owing to the high number of queries required for complex problems, particularly those rich in global constraints.

\paragraph{LLM‑driven CA and Bias‑free Acquisition.} Mechqrane \emph{et~al.}\ introduced \emph{ACQNOGOODS} and \emph{LLMACQ}, the first query‑based CA framework that (i) does \emph{not} require an explicit bias because it incrementally learns \emph{nogoods} rather than constraints and (ii) embeds a fine‑tuned BERT component that interprets natural‑language feedback, extracts the violated constraint from the user's explanation, and reduce the number of queries required to reach the target network\cite{mechqranellm}. Their empirical results show orders‑of‑magnitude query savings over QUACQ2 while handling arbitrary arity constraints.

\paragraph{Hybrid CA.} It combines passive and active learning. Typically, passive learning generates an initial model from a few examples, which are then actively refined \cite{balafashybrid}. This reduces the query load compared to pure active learning. However, if the initial passive phase produces over-fitted global constraints, the active phase may struggle or require many queries for correction \cite{bessiere2016new}. Existing hybrid methods lack dedicated mechanisms to specifically address and rectify over-fitting before proceeding to general refinement.

\paragraph{Contribution to the Context.} This work introduces a hybrid framework that distinguishes itself by incorporating a novel dedicated \emph{Query-Driven Interactive Refinement} phase (Phase 2) explicitly designed to tackle the over-fitting of global constraints inherited from the passive learning phase (Phase 1). 

\section{Methodology: Hybrid Acquisition with Interactive Refinement}
\label{sec:methodology}

We propose a three-phase hybrid Constraint Acquisition (CA) framework, hereafter referred to as \textbf{Hybrid CA with Refinement (HCAR)}, designed to effectively learn constraint models, particularly in scenarios involving complex global constraints where initial learning from limited data typically leads to over-fitting. This study extends previous research. It builds upon the two-phase hybrid CA architecture previously introduced \cite{balafashybrid}, and expands upon the concept of query-driven refinement for addressing over-fitting, which was initially explored for \texttt{AllDifferent} constraints in Balafas et al. \cite{lion19balafas}.

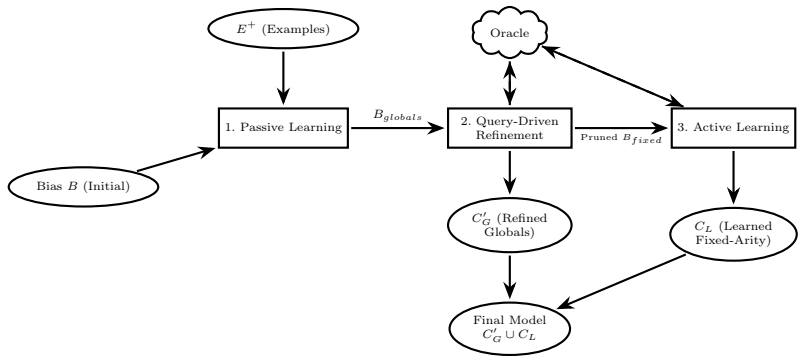
\begin{figure}[htbp] 
\centering
\begin{tikzpicture}[
    scale=0.65, transform shape,
    node distance=1.2cm and 1.5cm, 
    phase/.style={rectangle, draw, thick, minimum width=2.5cm, minimum height=0.8cm, align=center, font=\scriptsize},
    data/.style={ellipse, draw, thick, minimum width=2.2cm, minimum height=0.8cm, align=center, font=\scriptsize},
    oracle/.style={cloud, draw, thick, cloud puffs=10, cloud puff arc=125, aspect=1.8, minimum height=0.8cm, align=center, font=\scriptsize},
    arrow/.style={thick, ->, >=Stealth, shorten >=1pt, shorten <=1pt}, 
    dashed_arrow/.style={thick, ->, >=Stealth, dashed, shorten >=1pt, shorten <=1pt},
    label/.style={font=\tiny, midway} 
]

\node[data] (Eplus) {$E^+$ (Examples)};
\node[phase, below=1.2cm of Eplus] (Phase1) {1. Passive Learning}; 
\node[data, left=1.2cm of Phase1, yshift=-1.2cm] (Bias) {Bias $B$ (Initial)}; 

\node[phase, right=2.0cm of Phase1] (Phase2) {2. Query-Driven \\ Refinement}; 
\node[data, below=1cm of Phase2] (CGprime) {$C'_G$ (Refined \\ Globals)};
\node[oracle, above=1.0cm of Phase2] (Oracle) {Oracle}; 

\node[phase, right=2.0cm of Phase2] (Phase3) {3. Active Learning};
\node[data, below=1.2cm of Phase3] (CL) {$C_L$ (Learned \\ Fixed-Arity)};

\node[data, below=1cm of CGprime] (FinalModel) {Final Model \\ $C'_G \cup C_L$};

\draw[arrow] (Eplus) -- (Phase1);
\draw[arrow] (Bias) -- (Phase1);

\draw[arrow] (Phase1) -- node[above, font=\scriptsize, align=center] {$B_{globals}$} (Phase2);

\draw[arrow] (Oracle) -- (Phase2);
\draw[dashed_arrow] (Phase2) -- (Oracle);

\draw[arrow] (Phase2) -- (CGprime);
\draw[arrow] (Phase2) -- node[label, below, pos=0.5] {Pruned $B_{fixed}$} (Phase3);

\draw[arrow] (Oracle) -- (Phase3);
\draw[dashed_arrow] (Phase3) -- (Oracle);

\draw[arrow] (Phase3) -- (CL);
\draw[arrow] (CGprime) -- (FinalModel);
\draw[arrow] (CL) -- (FinalModel);

\end{tikzpicture}
\caption{High-level workflow of the three-phase hybrid CA framework.}
\label{fig:framework_workflow}
\end{figure}

Our framework generalizes and enhances this refinement approach, enabling it to handle a broader range of global constraints (\texttt{AllDifferent}, \texttt{Sum}, \texttt{Count}) and incorporate new mechanisms such as machine learning-based prior estimation, adaptive budget management, and specialized subset exploration for rejected candidates. The aim of this study was to combine these elements synergistically. As shown in Figure~\ref{fig:framework_workflow}, the HCAR  architecture unfolds in three distinct stages.
(1) Initial Passive Learning (Phase 1) generates a preliminary set of candidate global and fixed-arity constraints by analyzing a small number of positive examples.
(2) The Query-Driven Violation Refinement (Phase 2), which is the core of our approach, refines these global candidates. This phase employs oracle interaction, probabilistic confidence scoring, and subset exploration mechanisms to identify and address over-fitted constraints to improve their accuracy and scope.
(3) Final Active Learning (Phase 3) subsequently utilizes an established active learning algorithm initialized with refined global constraints and passively filtered fixed-arity candidates, to systematically determine any remaining fixed-arity constraints and ensure overall model completeness.

\subsection{Phase 1: Passive Candidate Generation and Initial Bias Refinement}
\label{subsec:phase1_passive}
The acquisition process commences with a passive learning phase that analyzes a typically small set of positive examples ($E^+$), which are complete assignments known to satisfy the unknown target model $\CT$. This phase has two primary objectives:
\begin{enumerate*}[label=(\alph*)]
    \item To efficiently generate an initial set of candidate global constraints ($B_{globals}$) by identifying the structural regularities within $E^+$.
    \item To construct and refine an initial bias set of fixed-arity constraints ($B_{fixed}$) by removing candidates inconsistent with $E^+$.
\end{enumerate*}

\paragraph{Generating Candidate Global Constraints ($B_{globals}$):}
This part employs techniques analogous to pattern-based methods such as ModelSeeker \cite{BelSim} and as detailed in our prior work \cite{balafashybrid}:
\begin{itemize}
    \item \textbf{Structural Pattern Detection:} The system examines the structure and values within examples in $E^+$. Problems with inherent grid or matrix structures (common in domains such as scheduling, rostering, or puzzles) \cite{flener2001matrix}, it specifically searches for consistent regularities along rows, columns, predefined blocks, or diagonals. For problems with list-based or less structured variable sets, it analyzes sequences and value groupings.
    \item \textbf{Domain Knowledge Input (optional):} Pre-existing knowledge about the problem structure can be incorporated. This might include defining groups of variables that are expected to interact (e.g., tasks in the same project, resources of the same type) or known capacities/limits that can guide parameter inference.
    \item \textbf{Constraint Matching and Instantiation:} Detected regularities are matched against a library $\Gamma_{global}$ of known global constraint types (e.g., \texttt{AllDifferent}, \texttt{Sum}, \texttt{Count}). If a pattern strongly suggests a particular global constraint type and is consistently observed across all examples in $E^+$, a candidate global constraint of that type is instantiated over the variables involved in the pattern (its scope). These form the set $B_{globals} = \{c_g^1, c_g^2, \dots, c_g^p\}$.
    \item \textbf{Parameter Inference:} For parameterized global constraints (e.g., the bound $b$ in $\texttt{Sum}(S, \bowtie, b)$ or the count $k$ in $\texttt{Count}(S, v, \bowtie, k)$), the system infers parameter values that are consistently satisfied by all examples in $E^+$. For instance, if the sum of variables in scope $S$ is always exactly 15 across all examples, a candidate $\texttt{Sum}(S, =, 15)$ might be generated. Domain knowledge can also provide or constrain these parameters.
    \item \textbf{Consistency Requirement:} Every generated global candidate $c_g \in B_{globals}$ must be satisfied by all provided positive examples $e \in E^+$.
\end{itemize}

\paragraph{Constructing and refining fixed-arity constraint bias ($B_{fixed}$):}
Concurrently, the system constructs an initial bias set $B_{fixed}$ containing all possible fixed-arity constraints (typically binary constraints in this study) derived from a basic relational language $\Gamma_{fixed}$ (e.g., using relations $\{\neq, =, >, <, \ge, \le\}$ over pairs of variables). This initial, potentially very large, set $B_{fixed}$ is immediately refined.
 Each constraint $c_f \in B_{fixed}$ is checked for consistency against all positive examples in $E^+$. If $c_f$ is violated by at least one example $e \in E^+$, it is removed from $B_{fixed}$.
The result of this step is a pruned set $B_{fixed}$ containing only those fixed-arity constraints that are consistent with all observed positive examples.

\paragraph{Output and Rationale for Subsequent Refinement:}
The output of Phase 1 consists of two sets:
\begin{enumerate}
    \item The set of candidate global constraints, $B_{globals}$.
    \item Refined set of candidate fixed-arity constraints $B_{fixed}$.
\end{enumerate}
While both $B_{globals}$ and $B_{fixed}$ are consistent with $E^+$, the limited nature of $E^+$ means these sets can still suffer from over-fitting. Constraints in $B_{globals}$ might correctly capture a pattern type but over-generalize its scope or parameters. Constraints in $B_{fixed}$ might represent coincidental relationships present in the few examples, but not true for the general problem.

\textbf{Example (Over-fitting in $B_{globals}$):} Consider learning constraints for a 4x4 Sudoku from only two positive examples, where variables are denoted as $x_{i,j}$ for row $i$ and column $j$. Phase 1 would likely correctly identify all row, column, and 2x2 block \texttt{AllDifferent} constraints and add them to $B_{globals}$. However, if \textbf{both} positive examples happen to also have distinct values along the main diagonal, Phase 1 might erroneously generate an additional candidate: $\texttt{AllDifferent}(\{x_{1,1}, x_{2,2}, x_{3,3}, x_{4,4}\})$. This diagonal constraint is consistent with the limited data but is not part of the true Sudoku rules – it is an over-fitted candidate in $B_{globals}$. Similarly, if the variables for the first two cells of the first row, $x_{1,1}$ and $x_{1,2}$, happen to have the value 1 in both examples, the unary constraints $x_{1,1}=1$ and $x_{1,2}=1$ would also be retained in the refined $B_{fixed}$. This occurs because these candidate constraints are satisfied by all provided positive examples and are therefore not filtered out, even though they are not true constraints of the general Sudoku problem.
The set $B_{globals}$ forms the primary input for the crucial \textbf{Query-Driven Interactive Refinement (Phase 2)}, which is designed to specifically address over-fitting in these complex global constraints. The refined fixed-arity bias $B_{fixed}$ is typically passed to the final \textbf{Active Learning phase (Phase 3)} for further investigation if needed.

\subsection{Phase 2: Query-Driven Interactive Refinement}
\label{subsec:phase2_refinement}
This phase is the core innovation of our framework, specifically designed to address the over-fitting issues often present in the initial set of candidate global constraints $B_{globals}$ generated from limited data in Phase 1. It takes $B_{globals}$, interacts with an oracle $O$, and produces a refined set of validated global constraints $C'_G$ (which will form part of the final learned model $\CL$). A high-level overview of how this interactive refinement system operates is presented in Algorithm~\ref{alg:hybrid_ca}.

\begin{algorithm}
\caption{Hybrid CA with Query-Driven Violation}
\label{alg:hybrid_ca}
\begin{algorithmic}[1]
\Require
    Positive examples $E^+$, Oracle $O$, 
    total budget $Budget_{total}$, time limit $T_{max}$, 
    confidence thresholds $\theta_{min}, \theta_{max}$, 
    Bayesian noise $\alpha$, max subset depth $d_{max}$
\Ensure
    Final learned constraint network $\CLfinal$

\Statex \textbf{// Phase 1: Passive Candidate Generation}
\State $(B_\text{globals},\;B_{\text{fixed}}) \gets
       \text{ExtractConstraintCandidates}(E^+)$
      \State $C'_G \gets \emptyset$, $Q_{used} \gets 0$, $Budget_{pool} \gets 0$

\ForAll{$c \in B_\text{globals}$}
    \State $P(c) \gets \text{MLPrior}(c)$, $Level(c) \gets 0$
    \State $B_{alloc}(c) \gets Budget_{total} / |B_\text{globals}|$ \Comment{Uniform initial budget allocation}
\EndFor

\Statex \textbf{// Phase 2: Query-Driven Interactive Refinement}
\While{$
    B_{\mathrm{globals}}\neq\emptyset
    \;\land\;
    Q_{\mathrm{used}}<Budget_{\mathrm{total}}
    \;\land\;
    \text{time}<T_{\max}
$}
    \State Select $c \in B_\text{globals}$ with lowest $P(c)$ and $B_{alloc}(c) > 0$
    \State $q_c \gets 0$
\While{$
   \theta_{\min} < P(c) < \theta_{\max}
   \;\land\;
   q_c < B_{\mathrm{alloc}}(c)
$}

        \State $Y \gets$ \textbf{GenerateQuery}($c$, $C'_G$, $B_\text{globals} \setminus \{c\}$) \Comment{Defined in Sec 4.2.3}
        \If{$Y$ is null}
            \State $P(c) \gets \theta_{max}$ 
            \State \textbf{break} 
        \EndIf
        \State $R \gets \text{AskOracle}(O, Y)$
        \If{$R$ = \textit{"Valid"}}
            \State $P(c) \gets 0$ \Comment{Constraint refuted}
            \State \textbf{break}
        \Else
            \State $P(c) \gets$ \textbf{UpdateBayes}($P(c)$, $Y$, $R$, $c$, $\alpha$)
        \EndIf
        \State $q_c \gets q_c + 1$, $Q_{used} \gets Q_{used} + 1$
    \EndWhile

    \If{$P(c) \ge \theta_{max}$}
        \State Move $c$ to $C'_G$, remove $c$ and its descendants from $B_\text{globals}$
        \State $Budget_{pool} \gets Budget_{pool} + B_{alloc}(c) - q_c$
         \If{$Budget_{pool} > 0$}
            \State Redistribute $Budget_{pool}$ evenly to remaining $B_\text{globals}$
            \State $Budget_{pool} \gets 0$
        \EndIf
    \ElsIf{$P(c) \le \theta_{min}$}
        \State Remove $c$ from $B_\text{globals}$
        \If{$Level(c) < d_{max}$}
            \State $S \gets$ \textbf{GenerateSubsets}($c$) \Comment{First, middle, last variable removal}
            \ForAll{unseen $s \in S$ with $|\text{vars}(s)| \ge 2$}
                \State $P(s) \gets \text{MLPrior}(s)$, $Level(s) \gets Level(c)+1$
                \State Allocate $B_{alloc}(s) \gets$ budget share from $B_{alloc}(c)$
                \State Add $s$ to $B_\text{globals}$
            \EndFor
        \EndIf
    \EndIf
\EndWhile

\Statex \textbf{// Phase 3: Final Active Learning}
\State $C_L \gets \text{MQuAcq-2}\!\bigl(O,\; C'_G,\; B_{\text{fixed}}\bigr)$
\State $\CLfinal \gets C'_G \cup C_L$
\State \Return $\CLfinal$

\end{algorithmic}
\end{algorithm}

The interactive refinement process is built upon several interacting components:
\begin{itemize}
    \item \textbf{Probabilistic Confidence Management:} Each constraint $c$ in the working bias set (initially populated from $B_{globals}$) is associated with a confidence score $P(c)$ representing the system's belief in its validity. This score is initialized using priors from a pre-trained machine learning model and then dynamically updated based on oracle feedback.
    \item \textbf{Machine Learning-Informed Priors:} An offline-trained classifier provides initial confidence estimates for constraints based on their structural and statistical features. This helps to prioritize early investigative efforts.
    \item \textbf{Iterative Query-Driven Violation Loop:} The system iteratively selects a constraint for investigation. A dedicated sub-process then generates informative queries specifically designed to test this constraint. Oracle's responses to these queries lead to updates in the constraint's confidence score. This sub-process is an integral part of the main loop in Algorithm~\ref{alg:hybrid_ca} (lines 8-18).
    \item \textbf{Bayesian Confidence Updates:} After each oracle response, the confidence score of the investigated constraint is updated using Bayes' theorem, allowing for a principled aggregation of evidence. If a constraint clearly contradicts an example, this leads to its immediate refutation.
    \item \textbf{Decision Framework with Structured Pruning:} Based on whether a constraint's confidence crosses predefined acceptance ($\theta_{max}$) or rejection ($\theta_{min}$) thresholds, it is either accepted into $C'_G$ or rejected from the bias.
    \item \textbf{Specialized Subset Exploration:} If a global constraint is rejected, a mechanism attempts to recover valid substructures. This typically involves generating subconstraints by removing single variables from strategic positions (first, middle, last) while preserving the original constraint type and parameters. These new subset candidates are then added to the bias for investigation.
    \item \textbf{Adaptive Query Budget Allocation:} The limited resource of oracle queries is managed dynamically. An initial budget is allocated to each constraint, and this budget is consumed during its investigation. Unused budget from accepted constraints is pooled and redistributed, while rejected constraints pass their remaining budget to any subsets they generate.
\end{itemize}
These components work in collaboration to systematically examine the initial global constraint candidates, filter out over-fitted ones, and through targeted analysis including subset exploration, establish more accurate scopes for global constraints, ultimately building a high-confidence set $C'_G$. The detailed mechanisms are elaborated in the following subsections.

\subsubsection{Probabilistic Confidence Framework}
\label{subsubsec:phase2_confidence}
To manage the inherent uncertainty about the validity of constraints derived from limited data, we employ a probabilistic approach. Each constraint $c$ in the bias set $B_{globals}$ is associated with a confidence score $P(c) \in [0,1]$. This score represents the system's current estimate of the probability that $c$ is part of the true target constraint network $\CT$. These scores are not static; they are dynamically updated based on the evidence obtained from interactions with the oracle $O$.

The framework uses two thresholds to guide the decision-making process:
\begin{itemize}
    \item \textbf{Acceptance Threshold ($\theta_{max}$):} Typically set close to 1 (e.g., 0.9 or 0.95). If evidence accumulates such that $P(c) \ge \theta_{max}$, the system concludes with high confidence that $c$ is a valid constraint. It is then moved from the bias $B_{globals}$ to the set of learned constraints $C'_G$.
    \item \textbf{Rejection Threshold ($\theta_{min}$):} Typically set close to 0 (e.g., 0.1 or 0.05). If evidence refutes the constraint such that $P(c) \le \theta_{min}$, the system concludes with high confidence that $c$ is invalid or over-fitted. It is then removed from the bias $B_{globals}$. Further actions depend on whether subset exploration is applicable (Section~\ref{subsubsec:phase2_subset_exploration}).
\end{itemize}
Constraints with confidence levels between these thresholds ($\theta_{min} < P(c) < \theta_{max}$) are considered undecided and remain in the bias set for potential further investigation, subject to resource limitations (query budget, time). This probabilistic approach allows the system to incrementally refine its beliefs and make principled decisions even when absolute certainty is unattainable.

\subsubsection{Machine Learning Model for Prior Confidence Estimation}
\label{subsubsec:phase2_ml_model}

Instead of initializing all constraints with a uniform prior confidence (e.g., $P(c)=0.5$), we leverage machine learning to provide more informed starting points. An XGBoost classifier is trained offline on a purpose-built dataset to predict the likelihood of a constraint being valid based purely on its structural and statistical properties.

\paragraph{Model Training and Data:} The training data is generated independently of the target problem being learned to ensure the model learns general structural features rather than problem-specific ones. It consists of labeled constraint instances derived from a diverse set of known CSP benchmarks, primarily from CSPLib \cite{csplib}. A detailed list of the problems used for training is provided in Appendix~\ref{app:csplib_problems}. This set includes representative problems from domains such as Quasigroup completion, N-Queens puzzles, Magic Squares, Number Partitioning, and scheduling problems such as Car Sequencing. For each benchmark, its ground-truth constraints are labeled 'valid'. Invalid constraints are generated by simulating the passive learning phase on small subsets of solutions from these benchmarks, capturing constraints that arise due to sampling artifacts (over-fitting). Additionally, invalid examples are created by synthetically perturbing valid constraints; for example, by slightly altering the scope of a valid global constraint (e.g., adding/removing a variable at random), changing its numerical bound, or modifying the relation type. These perturbations aim to create "near-miss" invalid constraints that are structurally similar to valid ones, helping the model learn finer distinctions. This creates a dataset where the model learns to differentiate structurally correct constraints from common over-fitting patterns.

\begin{table}[H]
\centering
\caption{Features extracted for ML-based prior estimation (per constraint $c$). Features 9-14 computed per inferred dimension $d$. \cite{tsouros2024learning}}
\label{tab:ml_features}
\resizebox{\textwidth}{!}{%
\begin{tabular}{@{}clll@{}}
\toprule
\textbf{ID} & \textbf{Feature Name} & \textbf{Type} & \textbf{Description} \\
\midrule
1 & \texttt{relation} & String & Type name of the constraint (e.g., 'AllDifferent', 'Sum', 'Count', $\le$). \\
2 & \texttt{arity} & Int & Number of variables in scope. \\
3 & \texttt{has\_constant} & Bool & True if constraint involves a numerical constant. \\
4 & \texttt{constant\_value} & Int & Value of the numerical constant, if present (0 otherwise). \\
5 & \texttt{var\_name\_same} & Bool & True if all scope variables share the same base name. \\
6 & \texttt{var\_ndims\_same} & Bool & True if all scope variables have the same inferred number of dimensions. \\
7 & \texttt{var\_ndims\_max} & Int & Maximum inferred dimension count among scope variables. \\
8 & \texttt{var\_ndims\_min} & Int & Minimum inferred dimension count among scope variables. \\
\midrule
\multicolumn{4}{l}{\textit{Features 9-14 computed for each dimension $d$ (up to $\texttt{var\_ndims\_max}-1$):}} \\
\midrule
9 & \texttt{var\_dim\textit{d}\_has} & Bool & True if all relevant variables have an index for dimension $d$. \\
10 & \texttt{var\_dim\textit{d}\_same} & Bool & True if all index values in dimension $d$ are identical. \\
11 & \texttt{var\_dim\textit{d}\_max} & Int & Maximum index value in dimension $d$. \\
12 & \texttt{var\_dim\textit{d}\_min} & Int & Minimum index value in dimension $d$. \\
13 & \texttt{var\_dim\textit{d}\_avg} & Float & Average index value in dimension $d$. \\
14 & \texttt{var\_dim\textit{d}\_spread} & Float & Range (max - min) of index values in dimension $d$. \\
\bottomrule
\end{tabular}
}
\end{table}

\paragraph{Feature Engineering:}
Rather than designing an entirely new feature set, we adopted a set of well-established and tested features previously proposed in the CA literature, specifically those detailed by Tsouros et al. \cite{tsouros2024learning}. This approach leverages features already demonstrated to be informative for distinguishing valid constraints from spurious ones based on their structural properties. As detailed in Table~\ref{tab:ml_features}, these features include the constraint's type (\texttt{relation}), arity, the presence and value of any constants (\texttt{has\_constant}, \texttt{constant\_value}), the consistency of base names among variables in the scope (\texttt{var\_name\_same}), and a set of statistics regarding the inferred dimensionality and index patterns of variables within the constraint's scope (e.g., min/max/average/spread of indices per dimension, denoted as \texttt{var\_dim\textit{d}\_*} features). This feature set provides a multi-faceted quantitative descriptor for each constraint, enabling the classifier to discern patterns indicative of validity or over-fitting.

\paragraph{Initialization Procedure:} At the start of Phase 2, for every candidate $c$ in $B_{globals}$, its features are extracted, processed, and fed into the pre-trained XGBoost model. The model's output probability (predicting the likelihood of the 'valid' class) is used as the initial confidence score $P(c)$, as detailed in Algorithm~\ref{alg:hybrid_ca} (line 4). 


\subsubsection{Query-Driven Refinement Loop}
\label{subsubsec:phase2_loop}
The core of Phase 2 is the query-driven refinement loop, explicitly initiated in Algorithm~\ref{alg:hybrid_ca} (line 76), which details its operational flow. This loop continues as long as the working bias set $B_{globals}$ is non-empty and overall resource limits (total queries $Budget_{total}$, global time $T_{max}$) are not exceeded, as specified by the loop's continuation condition (Algorithm~\ref{alg:hybrid_ca}, line 6). Each iteration performs the following steps:

\begin{enumerate}
    \item \textbf{Candidate Constraint Selection (Algorithm~\ref{alg:hybrid_ca}, line 7):} A target constraint $c$ is selected from $B_{globals}$ for investigation. This selection is guided by a strategy that prioritizes constraints with lower current confidence $P(c)$ (indicating higher uncertainty or more refuting evidence) and sufficient allocated query budget $B_{alloc}(c)$ remaining for investigation. The selection may also consider the constraint's position in a subset exploration hierarchy, a concept that will be detailed in Section~\ref{subsubsec:phase2_subset_exploration} when discussing the handling of rejected constraints.

    \item \textbf{Query-Driven Violation for Selected Constraint $c$:}
      Once $c$ is selected, its validity is determined through an iterative process detailed in Algorithm~\ref{alg:hybrid_ca} (lines 9-17). This investigation operates within the specific budget $B_{alloc}(c)$ allocated to $c$ and the overall system limits. It interactively queries the oracle $O$ as follows:
        \begin{itemize}
       \item \textbf{Query Generation (\texttt{GenerateQuery}, Algorithm~1, line 10):} An assignment $Y$, which will be posed as a query to the oracle, is generated. This assignment is specifically designed to test the selected constraint $c$ within the context of already validated global constraints $C'_G$ and other constraints currently in the bias $B_{globals} \setminus \{c\}$. This query generation adapts PQ-GEN principles \cite{tsouros2023guided} and is formulated as an auxiliary CSP. Let $X_{rel}$ be the union of the following sets of variables: 1) the scope of $c$ ($\text{vars}(c)$), 2) the scopes of all constraints in $C'_G$ ($\bigcup_{c' \in C'_G} \text{vars}(c')$), and 3) the scopes of all constraints in $B_{globals} \setminus \{c\}$ ($\bigcup_{c'' \in B_{globals} \setminus \{c\}} \text{vars}(c'')$). The auxiliary CSP aims to find an assignment $Y$ over variables $X_{rel}$ such that:
    \begin{enumerate}[label=(\alph*)]
        \item $Y \models c'$ for all $c' \in C'_G$: the assignment $Y$ must satisfy all global constraints already validated and accepted into $C'_G$.
        \item $Y \models c''$ for all $c'' \in (B_{globals} \setminus \{c\})$: the assignment $Y$ must also satisfy all other candidate global constraints currently in the bias $B_{globals}$ (excluding the one being tested, $c$).
        \item $Y \not\models c$: the assignment $Y$ must violate the specific constraint $c$ that is currently under investigation.
        \item $Y \notin H_c$: the assignment $Y$ must be unseen for testing $c$, meaning it has not been previously generated and presented to the oracle in the context of evaluating $c$.
    \end{enumerate}
    This CSP is solved using a conventional constraint solver within a specified time limit. If no such assignment $Y$ can be found, `GenerateQuery` returns null. In this case, the constraint `c` is considered untestable or implied by the current model. Its confidence is set to the acceptance threshold, $P(c) \gets \theta_{max}$, to ensure it is accepted and removed from the pool of candidates, thereby preventing an infinite loop (Algorithm~1, lines 11-13).

            \item \textbf{Oracle Interaction:} If a query $Y$ is successfully generated, it is presented to the oracle $O$, which returns a response $R$ ('Valid' or 'Invalid') indicating if $Y$ is a true solution to the underlying problem.

            \item \textbf{Confidence Update:} The confidence score $P(c)$ is then updated based on the oracle's response $R$ and whether $Y$ satisfied or violated $c$, using the Bayesian update mechanism described in Section~\ref{subsubsec:phase2_bayes_update}. This process also handles immediate refutation if $R$ is 'Valid' and $Y$ violates $c$.

            \item \textbf{Iteration within Budget:} This query-generation, oracle-interaction, and confidence-update cycle for constraint $c$ repeats as long as its allocated budget $B_{alloc}(c)$ is not exhausted, its confidence $P(c)$ remains between the decision thresholds $\theta_{min}$ and $\theta_{max}$, and global resource limits are respected.
        \end{itemize}
    The query-driven violation sub-process concludes by returning the final confidence $P(c)$ for the investigated constraint and the total number of queries $Q_{used\_for\_c}$ consumed during its investigation.

    \item \textbf{Decision Making and State Update:} Based on the final $P(c)$ returned from the focused learning sub-process, the system applies the decision framework (Accept/Reject/Undecided, detailed in Section~\ref{subsubsec:phase2_decision}). The sets $C'_G$ (refined global constraints) and $B_{globals}$ (working bias) are updated accordingly. The total query count $Q_{total\_used}$ is incremented by $Q_{used\_for\_c}$, and the budget pool $Budget_{pool}$ (for redistribution) may be updated if $c$ was accepted.
\end{enumerate}

\subsubsection{Bayesian Confidence Updates}
\label{subsubsec:phase2_bayes_update}

This section describes how our framework updates the confidence in a constraint $c$ being part of the true target constraint network $\CT$. After presenting a query assignment $Y$ to the oracle, which responds with $R$ (`Valid' or `Invalid'), we adjust the confidence score $P(c)$, representing the probability $P(c \in \CT)$, using Bayesian reasoning.

Each constraint $c$ has a confidence score $P(c) \in [0,1]$, initially set by a machine learning model (Section~\ref{subsubsec:phase2_ml_model}). The oracle’s response to a query provides evidence to refine this score. We use Bayes’ theorem to compute the updated (posterior) probability:
\[
P(c \in \CT \mid E) = \frac{P(E \mid c \in \CT) \cdot P(c \in \CT)}{P(E)}
\]
Here, $E$ is the evidence from the oracle’s response $R$, $P(c \in \CT)$ is the prior confidence (before seeing $E$), and $P(E \mid c \in \CT)$ is the likelihood of observing $E$ if $c$ is valid. The denominator $P(E) = P(E \mid c \in \CT) P(c \in \CT) + P(E \mid c \notin \CT) P(c \notin \CT)$ normalizes the result.

 We use a noise parameter $\alpha = 0.42$, empirically tuned on validation problems to balance responsiveness to new evidence with stability. This parameter $\alpha$ models the uncertainty in a single query’s ability to confirm or refute $c$.

\paragraph{Case 1: Oracle Responds `Invalid' (Negative Example)}
If the oracle labels $Y$ as `Invalid', $Y$ is not a solution to $\CT$. As $Y$ violates $c$, this supports $c$ being part of $\CT$, as a true constraint should reject invalid assignments. We set:
\begin{itemize}
    \item $P(E = \text{`Invalid'} \mid c \in \CT) = 1 - \alpha$, reflecting high likelihood that a valid $c$ correctly forbids $Y$.
    \item $P(E = \text{`Invalid'} \mid c \notin \CT) = \alpha$, assuming a false $c$ is less likely to align with $Y$’s invalidity.
\end{itemize}
Applying Bayes’ theorem increases $P(c)$, strengthening our belief in $c$’s validity.

\paragraph{Case 2: Oracle Responds `Valid' (Positive Example)}
If the oracle labels $Y$ as `Valid', $Y$ is a solution to $\CT$. As $Y$ violates $c$, this contradicts $c$’s validity, as a true constraint must allow all valid solutions. In this case:
\begin{itemize}
    \item We set $P(c) = 0$, immediately rejecting $c$.
    \item $c$ is removed from the bias set $B_{globals}$.
    \item Bayesian calculations are skipped, as a single counterexample definitively refutes $c$.
\end{itemize}

\paragraph{Example}
Consider a constraint $c = \texttt{Sum}(x_1, x_2, x_3) = 12$ with initial $P(c) = 0.6$. We generate a query $Y: x_1 = 4, x_2 = 4, x_3 = 5$ (sum = 13, violating $c$). The oracle responds:
\begin{itemize}
    \item \textbf{`Invalid'}: $Y$ violates $c$ and is not a valid assignment, supporting $c$. Bayes’ update with $\alpha = 0.42$ increases $P(c)$, e.g., to $\approx 0.75$.
    \item \textbf{`Valid'}: $Y$ violates $c$ but it is a valid assignment, refuting $c$. We set $P(c) = 0$ and discard $c$.
\end{itemize}
\subsubsection{Decision Framework} 
\label{subsubsec:phase2_decision}
The final $P(c)$ after the query-driven violation loop determines the outcome:
\begin{enumerate}
    \item \textbf{Accept ($P(c) \ge \theta_{max}$):} The constraint $c$ is moved from the working bias $B_{globals}$ to the learned set of global constraints $C'_G$ (line 22). 
    The remaining allocated budget for $c$ is added to a redistribution pool. This pool is then immediately and uniformly redistributed among all remaining candidate constraints in $B_{globals}$ to fund further investigation (lines 23-26).
    \item \textbf{Reject ($P(c) \le \theta_{\min}$):} $c$ is removed from $B_{globals}$. Pruning only affects $c$ itself. If $c$ is suitable (as explained below), Subset Exploration (Section~\ref{subsubsec:phase2_subset_exploration}) is triggered, receiving $c$'s remaining budget.
    \item \textbf{Undecided ($\theta_{min} < P(c) < \theta_{max}$):} $c$ remains in $B_{globals}$  with its updated confidence and reduced budget, potentially revisited later.
\end{enumerate}

\subsubsection{Subset Exploration for Rejected Constraints}
\label{subsubsec:phase2_subset_exploration}

A critical challenge in learning from limited data is that constraints identified passively may be structurally correct in their type (e.g., a \texttt{Sum} relationship exists) and parameters (e.g., the numerical bound is meaningful), but incorrect in their scope (e.g., too many variables included). Simply discarding a constraint $c$ when its confidence $P(c)$ falls below the rejection threshold $\theta_{min}$ risks losing this potentially valuable partial information. To mitigate this knowledge loss, our framework incorporates a mechanism to explore subsets of rejected constraints, attempting to recover valid substructures. This exploration is managed within a dynamically evolving \textit{constraint hierarchy}. Given an initial candidate constraint $c$ which has been rejected, we consider that this constraint is at depth 0 (root) of the hierarchy. Subconstraints that are generated by removing variables from the scope of $c$ are at level 1. In this case, these constraints are \textit{siblings} and $c$ is their \textit{parent}, forming a parent-child lineage. Constraints generated from the children of $c$ are at level 2 and so on.

\paragraph{Triggering Conditions:}
This exploration mechanism is triggered specifically when a constraint $c$ is rejected ($P(c) \le \theta_{min}$) during the refinement process (Section~\ref{subsubsec:phase2_decision}). However, to prevent potentially tedious exploration down chains of increasingly smaller subsets (e.g., rejecting a subconstraint, generating its subconstraints, and so on), this mechanism is only invoked if the rejected constraint $c$ is not already too deep in the hierarchy generated by previous explorations. Specifically, its level $\text{Level}(c)$ 
must be less than a predefined maximum depth (e.g., 3). Also, to improve efficiency, if the rejected constraint $c$ is below level 1 (i.e. it is not the initial candidate) any sibling constraints of $c$ 
and their descendants are also removed from $B_{globals}$.

\paragraph{Candidate Generation via Targeted Variable Removal (Algorithm~\ref{alg:hybrid_ca}, line 25):}
If triggered, the system attempts to generate potentially valid subconstraints from the rejected parent constraint $c$. As defined in Section~\ref{sec:background}, these subconstraints will share $c$'s type and parameters but operate on a reduced scope. Instead of exhaustively exploring all $n$ possible single-variable removals (where $n$ is the arity of $c$), a heuristic is employed. This heuristic generates up to three new subconstraint candidates by removing variables from structurally indicative positions within the parent's ordered variable scope $V_c = (v_1, v_2, \ldots, v_n)$:
\begin{enumerate}
    \item A candidate subconstraint $c_{sub\_first}$ with scope $(v_2, \ldots, v_n)$ is generated by removing the \textbf{first} variable ($v_1$) from the scope of $c$.
        \item A candidate subconstraint $c_{sub\_middle}$ is generated by removing the \textbf{middle} variable ($v_{\lfloor n/2 \rfloor}$) from the scope of $c$, resulting in the scope:
    \[(v_1, \ldots, v_{\lfloor n/2 \rfloor - 1}, v_{\lfloor n/2 \rfloor + 1}, \ldots, v_n)\]
    \item A candidate subconstraint $c_{sub\_last}$ with scope $(v_1, \ldots, v_{n-1})$ is generated by removing the \textbf{last} variable ($v_n$) from the scope of $c$.
\end{enumerate}
This heuristic is pragmatic and computationally inexpensive. While it may not be optimal for all constraint types or variable orderings (e.g., if the truly superfluous variable in an incorrectly scoped constraint is not at an end or the exact middle), it offers a good balance between exploration and efficiency for common cases where an initial candidate constraint has been \textbf{over-scoped} (i.e., defined over too many variables compared to the true constraint). Future work could explore more sophisticated variable removal strategies, potentially guided by variable importance metrics or other structural properties of the constraints and variables.

Each generated candidate $c_{sub}$ \textbf{inherits the type and parameters} (e.g., the relation $\bowtie$ and bound $b$ for a \texttt{Sum}, the value $v$, relation $\bowtie$, and count $k$ for a \texttt{Count}) directly from the rejected parent constraint $c$. This design embodies the core hypothesis: the constraint's fundamental relation and associated parameters might be correct (having been inferred consistently from initial examples), while the scope might have been over-generalized.

For each generated candidate subconstraint $c_{sub}$ the scope size of $c_{sub}$ must remain at least 2 (i.e., $|\text{vars}(s)| \ge 2$). This minimum arity is chosen to allow for the formation of at least binary relationships, which are common, though many global constraints inherently require arity $\ge 3$ to be meaningful (e.g., \texttt{AllDifferent} over 2 variables is simply $v_1 \neq v_2$). Subconstraints failing this are discarded. $c_{sub}$ must not already be present in the current bias set $B_{globals}$ or the set of already learned constraints $C'_G$.
If the subconstraint has these properties:
        \begin{itemize}
            \item $c_{sub}$ is added to the bias set $B_{globals}$.
            \item It is assigned an initial confidence score $P(c_{sub})$ from the ML model.
            \item It inherits a query budget. As described in Section~\ref{subsubsec:phase2_resources}, the total remaining budget of the rejected parent $c$ is distributed uniformly among all its valid and unseen children $c_{sub}$ generated in this step.
            \item $c$ is marked as the parent of $c_{sub}$, and the level is set, $\text{Level}(c_{sub}) = \text{Level}(c) + 1$.
        \end{itemize}

\paragraph{Special Pruning for Accepted \texttt{AllDifferent} Subsets:}
An additional rule governs the pruning process when dealing with subsets derived from a rejected \texttt{AllDifferent} constraint. If any one of the generated \texttt{AllDifferent} subset candidates (say $c_{sub}$) is subsequently \textbf{accepted} ($P(c_{sub}) \ge \theta_{max}$), the system performs a more aggressive pruning action. In addition to standard pruning, it also \textbf{immediately removes all sibling subsets} of $c_{sub}$ and their descendants from $B_{globals}$. This heuristic is based on the assumption that for \texttt{AllDifferent} constraints, there is often a single, maximal correct scope. Suppose a subset with a reduced scope is validated. In that case, it is considered less likely that other subsets (formed by removing different single variables from the same over-scoped parent) represent distinct, non-overlapping true constraints. Instead, they are more likely to be either incorrect or subsumed by the accepted subset, or would lead to redundant, less general true constraints. This aggressive pruning aims to accelerate convergence by avoiding potentially unnecessary exploration of closely related \texttt{AllDifferent} variations. While this heuristic might be overly aggressive in rare cases where multiple, distinct \texttt{AllDifferent} sub-scopes are valid and desirable, it has shown practical benefits in common scenarios.

Revisiting the \textbf{Illustrative Example}: Phase 1 yields the over-fitted constraint $c = \texttt{Sum}(\{v_1, v_2, v_3, v_4, v_5\}, \le, 20)$, but the true constraint is $\CT = \{\texttt{Sum}(\{v_1, v_2, v_4, v_5\}, \le, 20)\}$. In Phase 2, oracle queries eventually lead to the rejection of $c$ (i.e., its confidence falls below $\theta_{min}$). The Subset Exploration mechanism is then triggered, generating new candidates, including the correct one: $c_{sub\_middle} = \texttt{Sum}(\{v_1, v_2, v_4, v_5\}, \le, 20)$ (assuming $v_3$ was the middle variable). This new candidate, $c_{sub\_middle}$, is added to the working bias set $B_{globals}$, where it is initialized with a confidence score from the ML model and assigned a portion of the rejected parent's budget.
In a later iteration, $c_{sub\_middle}$ is selected for investigation. The query-driven violation process generates queries specifically to test it. Since $c_{sub\_middle}$ is part of the true model, any query that violates it (e.g., an assignment where $v_1+v_2+v_4+v_5 > 20$) will be classified as 'Invalid' by the oracle. This consistent, confirming evidence causes the confidence of $c_{sub\_middle}$ to increase via Bayesian updates. Eventually, its confidence surpasses $\theta_{max}$, and it is accepted into the refined set $C'_G$. This demonstrates how the exploration mechanism successfully recovers the valid constraint structure while preserving the original, potentially meaningful, numerical bound.
\subsubsection{Resource Management: Query Budget Allocation and Constraint Hierarchy}
\label{subsubsec:phase2_resources}
The interactive refinement process consumes oracle queries, a valuable and often limited resource. To manage this effectively, a dynamic query budget allocation strategy is employed, potentially operating within an overall global query limit $Budget_{total}$ for the entire phase. 

The query budget $B_{alloc}(c)$ for each constraint $c$ is managed through several mechanisms:
\begin{itemize}
    \item \textbf{Initial Allocation:} Upon entering Phase 2, the specified $Budget_{total}$ is distributed \textbf{uniformly} among all constraints $c$ in the initial bias set $B_{globals}$, establishing their starting $B_{alloc}(c)$.
    \item \textbf{Consumption:} During the focused investigation of a constraint $c$, $Q_{used\_c}$ queries are consumed, and its allocated budget is reduced accordingly: $B_{alloc}(c) \leftarrow B_{alloc}(c) - Q_{used\_c}$.
    \item \textbf{Reclaiming upon Acceptance:} If constraint $c$ is accepted into the learned set $C'_G$, its remaining budget, $B_{alloc}(c) - Q_{used\_c}$, is reclaimed and added to a central redistribution pool, $Budget_{pool}$.
    \item \textbf{Inheritance for Subsets:} If constraint $c$ is rejected and subsequently generates a set of subconstraint candidates $S_{sub}$, the remaining budget of $c$, $B_{alloc}(c) - Q_{used\_c}$, is \textbf{uniformly} distributed among these new subconstraints $s \in S_{sub}$ to fund their initial investigation.
    \item \textbf{Redistribution upon Acceptance:} 
    When a constraint $c$ is accepted into the learned set $C'_G$, its remaining budget is immediately reclaimed and redistributed. This reclaimed amount is uniformly distributed among all other constraints currently remaining in the bias set $B_{globals}$ 
    (lines 23-26).
\end{itemize}
This adaptive and hierarchical budget management system ensures that query resources are dynamically channeled to  constraints requiring more evidence, 
while resources are conserved from constraints that are resolved quickly. 
While uniform distribution is used here for simplicity, future work could explore more adaptive or prioritized budget allocation strategies based on constraint features or confidence levels.

\subsubsection{Termination Criteria}
\label{subsubsec:phase2_termination}
Phase 2 terminates when the bias set $B_{globals}$ is empty, or the global query budget $Budget_{total}$ is exhausted, or the time limit $T_{global}$ is reached. The output is the refined set $C'_G$.

\subsection{Phase 3: Final Active Learning Refinement}
\label{subsec:phase3_active}

Following the Query-Driven Interactive Refinement of global constraints in Phase 2, which produces a refined, high-confidence set of global constraints $C'_G$, the HCAR framework proceeds to a final active learning phase. The primary objective of this phase is to complete and finalize the model by learning or discarding the remaining candidate fixed-arity constraints. 

Phase 3 is initialized with two components:
\begin{enumerate}
    \item \textbf{Learned Global Constraints ($C'_G$):} The set of global constraints validated during Phase 2. These are treated as known and correct components of the model being built.
    \item \textbf{Candidate Fixed-Arity Constraints:} The bias for this phase is initialized with the set of candidate fixed-arity constraints $B_{fixed}$ that was generated and refined during Phase 1 (as described in Section~\ref{subsec:phase1_passive}). This set $B_{fixed}$ contains all binary (or other low-arity) constraints consistent with the initial positive examples $E^+$.
\end{enumerate}
The active learning algorithm in Phase 3 then operates on the full set of problem variables $X$ and their domains $R$, aiming to resolve the status of constraints in $B_{fixed}$ in the context of the already learned global constraints $C'_G$.

This phase employs a standard active Constraint Acquisition algorithm. The novelty in this part of HCAR is not in the active learning algorithm itself but in its deployment after the specialized global constraint refinement of Phase 2, using the outputs $C'_G$ and $B_{fixed}$ as its starting point.  In our implementation, we utilize MQuAcq-2 \cite{tsourosstructure}, which is known for its efficiency in learning multiple constraints from negative examples and its ability to exploit problem structure. The chosen active learning algorithm proceeds iteratively:
\begin{enumerate}
    \item \textbf{Query Generation:} The algorithm generates an informative membership query $Y$. This assignment is constructed to satisfy all constraints currently in its learned set (which starts as $C'_G$) but aims to violate at least one constraint from its working bias $B_{P3}$.
    \item \textbf{Oracle Interaction:} The query $Y$ is presented to the oracle $O$, which classifies it as 'Valid' or 'Invalid' with respect to the true target model $\CT$.
    \item \textbf{Model Update:}
        \begin{itemize}
            \item If $R$ is 'Valid': Constraints in $B_{P3}$ that were violated by $Y$ are removed from $B_{P3}$.
            \item If $R$ is 'Invalid': The algorithm typically employs procedures (e.g., variations of FindScope \cite{bessiere2013constraint} or FindC \cite{bessiere2013constraint}) to identify the specific constraint(s) from $\CT$ (and presumed to be in the original universal bias from which $B_{fixed}$ was derived) responsible for the violation. These newly identified fixed-arity constraints are added to the set of learned fixed-arity constraints and removed from $B_{fixed}$.
        \end{itemize}
\end{enumerate}
This cycle of query generation, oracle interaction, and model update continues.

Phase 3 ensures that simpler relationships, primarily fixed-arity constraints that were pre-filtered in Phase 1 into $B_{fixed}$, are now actively investigated and correctly classified (learned or discarded).

For instance, while Phase 2 might correctly learn a complex $\texttt{AllDifferent}$ over a set of tasks, Phase 3 would take the $B_{fixed}$ (which might contain $task_i < task_j$ if this held in all initial 5 examples) and actively query to confirm if this specific binary precedence is indeed a true constraint.

Phase 3 terminates when standard active CA convergence criteria are met. This typically occurs when:
\begin{itemize}
    \item The working bias $B_{fixed}$ becomes empty.
    \item The query generation mechanism can no longer produce an informative query that violates at least one of the remaining candidates in $B_{fixed}$ from the currently learned model. 
    
\end{itemize}

The output of Phase 3 is the set of learned fixed-arity constraints $C_L$. The final, complete model learned by the HCAR framework is the union $\CLfinal = C'_G \cup C_L$. While the overall model accuracy (e.g., solution-space precision and recall) is evaluated on the complete model $\CLfinal$, our experimental analysis will place particular emphasis on the composition and correctness of the refined global constraint set $C'_G$, as this is the primary output of our novel refinement phase.
\section{Experimental Setup}
\label{sec:experimental_setup}

We conduct an empirical evaluation to assess the effectiveness and efficiency of HCAR. The experiments are designed to:
\begin{enumerate*}[label=(\arabic*)]
    \item Quantify the ability of HCAR, particularly its Phase 2, to handle over-fitting in global constraints and learn accurate models (RQ1).
    \item Evaluate the query efficiency and computational performance of HCAR compared to alternative CA strategies (RQ2).
\end{enumerate*}

\subsection{Benchmark Problems}
\label{subsec:benchmarks}
We selected five benchmark problems from different domains to create a diverse and challenging test suite. These problems involve various types and combinations of global constraints, variable structures, and levels of complexity. For each benchmark, we defined a ground-truth constraint model ($\CT$) containing the set of constraints considered correct and complete for a specific instance. The characteristics of these instances are summarized below, with full formal descriptions provided in Appendix A.

\begin{enumerate}
    \item \textbf{Sudoku (9x9):} A classic logic puzzle involving $|\CT| = 27$ \texttt{AllDifferent} constraints on rows, columns, and 3x3 blocks.
    \item \textbf{UEFA Champions League Scheduling:} A sports scheduling problem assigning 32 teams into 8 groups, subject to cardinality (\texttt{Count}), tier separation (\texttt{AllDifferent}), and country separation constraints. $|\CT| = 19$ constraints.
    \item \textbf{Cloud VM Allocation:} A resource allocation problem assigning Virtual Machines (VMs) to Physical Machines (PMs), considering resource capacities (\texttt{Sum} inequalities), high availability rules, etc. $|\CT| = 72$ constraints.
    \item \textbf{University Exam Timetabling:} Assigning exams to timeslots, respecting no-clash, room capacity (\texttt{Sum} inequalities), and other constraints. $|\CT| = 24$ constraints.
    \item \textbf{Nurse Rostering:} Assigning nurses to shifts, satisfying coverage (\texttt{Count}/\texttt{Sum}), workload balance (\texttt{Sum} inequalities), and sequential constraints. $|\CT| = 21$ constraints.
\end{enumerate}

\subsection{Evaluated Methods}
\label{subsec:evaluated_methods}
We evaluate our proposed \textbf{HCAR} framework against a baseline and conduct internal ablations to assess the contribution of its components. All methods that include a passive learning phase learn from an initial set of 5 positive examples.  We later expand on this by presenting results where the size of this initial example set is increased to analyze how performance scales with data availability.

\begin{enumerate}
    \item \textbf{HCAR (Proposed Method):} Our three-phase framework:
        \begin{itemize}
            \item Phase 1 (Passive Learning): Generates $B_{globals}$ from 5 positive examples.
            \item \textbf{HCAR (Proposed Method):} Our three-phase framework. The hyperparameters for the refinement phase were set as follows, based on a combination of standard practices and preliminary tuning on a separate validation set (not used in the final evaluation):
    \begin{itemize}
        \item \textbf{Decision Thresholds:} The confidence thresholds $\theta_{max}=0.9$ and $\theta_{min}=0.1$ were chosen as they represent common, stringent values for acceptance and rejection in probabilistic systems, requiring a high degree of certainty before a decision is made.
        \item \textbf{Bayesian Noise:} The noise parameter $\alpha=0.42$ was determined through empirical tuning on a set of validation problems (distinct from our test benchmarks). This value was found to provide a good balance between responsiveness to new evidence from oracle queries and stability against noisy or misleading individual queries.
        \item \textbf{Subset Exploration Limit:} The maximum subset depth was set to 3 as a pragmatic heuristic to prevent excessive exploration of increasingly niche sub-problems and to limit combinatorial explosion, while still allowing for meaningful recovery of over-scoped constraints.
        \item \textbf{Resource Limits:} The global query budget for Phase 2 ($Budget_{total}=1500$) and the time limit ($T_{global}=600$s), along with the per-query solver time limit ($t_{limit}=5$s), were set to values sufficiently large to allow the algorithm to converge naturally on our benchmark problems. They serve as a safeguard against pathological cases rather than a binding constraint on performance in these experiments.
    \end{itemize}
            \item Phase 3 (Final Active Learning): MQuAcq-2 \cite{tsouros2018efficient} initialized with $C'_G$ from Phase 2. MQuAcq-2 is a complete algorithm that runs until convergence of $B_{fixed}$. There is no query budget or time limit.
        \end{itemize}

    \item \textbf{Baseline: Two-Phase Hybrid (HCAR-NoRefine)}
        \begin{itemize}
            \item Phase 1 (Passive Learning): Same as HCAR, generating $B_{globals}$ from 5 positive examples.
            \item Phase 2: Skipped. The $B_{globals}$ are directly passed as the initial learned set.
            \item Phase 3 (Active Learning): MQuAcq-2 \cite{tsouros2018efficient} is used to complete the model, starting with $B_{globals}$. This baseline directly shows the impact of omitting our refinement phase.
        \end{itemize}

    \item \textbf{Ablation Studies for HCAR's Phase 2:} To understand the contribution of individual components within our refinement phase, we evaluate variants of HCAR where specific mechanisms are disabled:
        \begin{itemize}
            \item \textbf{HCAR (No ML Priors):} Phase 2 initializes all $P(c)$ to a flat 0.5 instead of using the XGBoost model predictions.
            \item \textbf{HCAR (No SVR):} Phase 2 proceeds as usual, but the  Subset Exploration mechanism (Section~\ref{subsubsec:phase2_subset_exploration}) is disabled. Rejected constraints are simply removed from the bias.
            \item \textbf{HCAR (No Bayesian Update):} This variant replaces the Bayesian framework with a one-shot decision rule. For each candidate constraint, the system generates a single query that violates it. The decision is then immediate and final:
            \begin{itemize}
                \item If the oracle's response is \textbf{'Valid'}, the constraint is rejected.
                \item If the response is \textbf{'Invalid'}, or if no such query can be generated, the constraint is accepted.
            \end{itemize}
        \end{itemize}
\end{enumerate}
\subsection{Implementation Details}
\label{subsec:implementation}

The proposed Hybrid Constraint Acquisition with Refinement (HCAR) framework, along with its HCAR-NoRefine variant used for baseline comparison, was implemented entirely in Python 3. For constraint modeling, and representation, the system leverages the \textbf{CPMpy} library \cite{guns2019increasing}, a Python-embedded modeling language that allows for high-level specification of constraint problems.

The task of solving CSPs and COPs for query generation within Phase 2 and the operations of the MQuAcq-2 algorithm in Phase 3 are using the \textbf{Google OR-Tools} CP\_SAT solver\cite{ortools}. OR-Tools provides an efficient suite of solvers for combinatorial optimization and constraint satisfaction.

The machine learning component responsible for estimating prior probabilities in Phase 2 (Section~\ref{subsubsec:phase2_ml_model}) employs an \textbf{XGBoost classifier} \cite{xgboost}. The implementation of this classifier, including its training and prediction functionalities, utilizes the popular \textbf{`scikit-learn`} library \cite{scikit-learn}. Model persistence (saving and loading the trained XGBoost model) is handled using `joblib`.

The MQuAcq-2 algorithm \cite{tsourosstructure}, which serves as the active learning engine in Phase 3 of HCAR and as the primary learner in the HCAR-NoRefine baseline, is implemented through the \textbf{PyConA} library (\url{https://github.com/CPMpy/PyConA}) \cite{pycona}. PyConA is a Python library dedicated to constraint acquisition, providing implementations of various active learning algorithms and utilities.

All experiments were conducted on a computer equipped with an Intel Core i7-8700 processor at 3.20 GHz and 16 GB of RAM, running Windows 10. The specific versions of Python and all libraries above were kept consistent throughout the experimental process to ensure the comparability and reproducibility of performance metrics across different configurations and benchmarks
\subsection{Evaluation Metrics}
\label{subsec:metrics}
To evaluate the HCAR framework and compare its performance against baselines and ablated versions, we employ a set of metrics that assess the accuracy of the learned constraint models at both the constraint and solution-space levels, the efficiency of the acquisition process in terms of queries posted and computational time, and the internal dynamics of the refinement phase.

Let $\CLfinal$ denote the final set of constraints learned by a given CA method, and $\CT$ be the ground-truth set of constraints for the benchmark problem.

\paragraph{Model Accuracy Metrics:}
These metrics evaluate the quality of the learned constraint set $\CLfinal$ relative to the target set $\CT$. The correctness of the solution space defined by $\CLfinal$ is assessed using solution-space precision and recall, which are standard evaluation measures in the constraint acquisition literature~\cite{prestwich2021classifier,kumar2022learning}.

\begin{itemize}
        \item \textbf{Learned Global Constraints ($\mathbf{|C'_G|}$):} The total count of global constraints in the refined set $C'_G$ produced by Phase 2. This is compared against the number of global constraints in the target model $\CT$ to assess how well the core structure was learned.

 \item \textbf{Solution-space Precision (S-Prec.):} This metric evaluates the correctness of the learned model. It is estimated by sampling solutions from $\CLfinal$ and calculating the fraction of them that are also valid solutions for the target model $\CT$:
        \[ \widehat{\text{S-Prec.}}(\CLfinal, \CT) = \frac{|\{s \in S_L : s \in \text{sol}(\CT)\}|}{|S_L|} \]
        where $S_L$ is a sample of $N_{samples}$ unique solutions generated by solving $\CLfinal$. A high S-Prec.\ value indicates that the learned model does not admit a significant number of invalid solutions (false positives).

\item \textbf{Solution-space Recall (S-Rec.):} This metric evaluates the coverage of the learned model. It is estimated by sampling solutions from the ground-truth model $\CT$ and calculating the fraction of them that are also accepted by the learned model $\CLfinal$:
        \[ \widehat{\text{S-Rec.}}(\CLfinal, \CT) = \frac{|\{s \in S_T : s \in \text{sol}(\CLfinal)\}|}{|S_T|} \]
        where $S_T$ is a sample of $N_{samples}$ unique solutions generated by solving $\CT$. A high S-Rec.\ value indicates that the learned model captures a large portion of the true solution space and does not erroneously reject valid solutions (false negatives).
\end{itemize}
For our experiments, solution-space precision and recall (S-Prec., S-Rec.) were estimated using a sample size of $N_{samples}=100$. To ensure a diverse and representative sample, these solutions were generated iteratively. After an initial solution was found, we repeatedly added constraints to the model to search for subsequent solutions, each required to have a \textbf{Hamming distance of at least 5} from all previously found solutions in the set\cite{balafas2024impact}. This process prevents solution clustering and provides a more robust test of the learned model across different regions of the solution space. Results are presented as percentages.

\paragraph{Efficiency Metrics:}
These metrics quantify the interactive burden and computational overhead of the CA process.
\begin{itemize}
    \item \textbf{Queries in Phase 2 ($Q_2$):} For HCAR and its ablations, the number of membership queries posed to the oracle exclusively during the Query-Driven Interactive Refinement phase (Phase 2).
    \item \textbf{Queries in Phase 3 ($Q_3$):} The number of membership queries posed during the final Active Learning phase (MQuAcq-2).
    \item \textbf{Total Queries ($Q_\Sigma$):} The aggregate number of membership queries ($Q_2 + Q_3$ for HCAR variants, or total MQuAcq-2 queries for baselines). This is the primary indicator of the overall burden on the domain expert or oracle.
    \item \textbf{Total Time (T (s)):} The total wall-clock time (in seconds) required for the completion of all learning phases of the CA method being evaluated (specifically Phase 2 and Phase 3 for HCAR).
\end{itemize} 

\paragraph{Refinement-Specific Internal Metrics (HCAR Phase 2):}
To provide deeper insights into the functioning and effectiveness of HCAR's Query-Driven Interactive Refinement phase, we track several internal metrics specifically for Phase 2 (as seen in Table~\ref{tab:refinement_stats_updated}):
\begin{itemize}
    \item \textbf{Constraints Removed:} The total number of candidate constraints eliminated from the bias during Phase 2. 
    A constraint can be removed either by being directly refuted by oracle feedback or by being proactively pruned when the acceptance of a related constraint (e.g., a more specific subset) makes it redundant.
    \item \textbf{Subsets Generated:} The total number of new subset constraints created by the Single Variable Removal mechanism.
    \item \textbf{Subsets Accepted:} The count of generated subset candidates that were  accepted into $C'_G$ during Phase 2.
    \item \textbf{Subsets Rejected:} The count of SVR-generated subset candidates that were rejected from the bias during Phase 2.
    \item \textbf{Average Queries per Accepted Subset (Avg. Q per Accept. Sub.):} The average number of oracle queries consumed during the query-driven violation sub-process specifically for those generated subset candidates that were accepted.
\end{itemize}

\section{Results and Analysis}
\label{sec:results}
This section presents an empirical evaluation of our proposed Hybrid CA with Refinement (HCAR) framework. We analyze its performance in learning accurate CSP models from a sparse initial set of examples, its efficiency in terms of oracle interaction and computational time, and the specific contributions of its core components, particularly the query-driven interactive refinement phase (Phase 2). The evaluation is conducted against a baseline (HCAR-NoRefine, which omits Phase 2) and through internal ablation studies to assess HCAR's capabilities and the impact of its design choices.

\subsection{Comparative Performance: HCAR versus Hybrid Without Refinement}
\label{subsec:results_baselines}

\begin{table}[htbp]
\small
\setlength{\tabcolsep}{4pt}     
\centering
\caption{Overall performance comparison. Five initial examples; identical query/time caps. }
\label{tab:overall_perf_nocprec}
\begin{tabular}{@{}l l r r r r r r r r@{}} 
\toprule
\textbf{Prob.} & \textbf{Alg.} & $\mathbf{|B_{globals}|}$ &
S-Prec. & S-Rec. & 
$\mathbf{|C'_G|}$ & $\mathbf{Q_2}$ & $\mathbf{Q_3}$ &
$\mathbf{Q_\Sigma}$ & \textbf{T (s)} \\
\midrule
Sudoku (9×9)   & HCAR          &  49 & \textbf{100\%} & \textbf{100\%} & \textbf{27} & 241 & 567 & 808 & 234.1 \\ 
               & NoRef         &  49 & 100\% & 82\% & 29 & - & 450 & 450 & 33.1 \\
\addlinespace
UEFA Sched.    & HCAR          &  27 & \textbf{100\%} & \textbf{100\%} & \textbf{19} & 233 &  18 & 251 & 48.9 \\ 
               & NoRef         &  27 & 100\% & 71\% & 23 & - & 61 & 61 & 41.3 \\
\addlinespace
Exam TT        & HCAR          &  46 & \textbf{100\%} & \textbf{100\%} & \textbf{24} & 108 & 64 & 172 & 69.4 \\
               & NoRef         &  46 & 100\% & 64\% & 31 & - & 124 & 124 & 28.7 \\
\addlinespace
VM Alloc.      & HCAR          &  118 & \textbf{100\%} & \textbf{100\%} & \textbf{77} &  284 & 67 & 351 & 18.9 \\
               & NoRef         &  118 & 100\% & 39\% & 97 & - & 145 & 145 & 29.3 \\
\addlinespace
Nurse Rost.    & HCAR          &  68 & \textbf{100\%}& \textbf{100\%} & \textbf{26} & 188 & 87 & 275  & 19.4 \\ 
               & NoRef         &  68 & 100\% & 33\% & 33 & - & 189 & 189 & 38.2 \\
\bottomrule
\end{tabular}

\footnotesize
S-Prec./S-Rec.\ = solution-space precision/recall;  
$|B_{globals}|$ = initial bias size of global constraints from Phase 1; $\mathbf{|C'_G|}$ = final global constraints learned (for HCAR, it is the number of constraints are after Phase 2; for NoRef, it is constraints after Phase 1);  
$Q_2$/$Q_3$ = queries in Phase 2/Phase 3;  
$Q_\Sigma$ = total queries.
\end{table}

We first evaluate the overall efficacy of HCAR against the HCAR-NoRefine baseline. Our initial and most challenging test case starts from a sparse dataset of only five positive examples. This choice is motivated because it simulates a realistic and difficult real-world scenario where obtaining a large set of verified solutions is often impractical or costly for a domain expert. Second, and more critically for our research, this data-limited condition is precisely where passive learning methods are most susceptible to over-fitting—learning spurious constraints from coincidental patterns in the few available examples. 
The impact of providing more initial examples is explored later in Section~\ref{subsec:results_more_examples}.

Table~\ref{tab:overall_perf_nocprec} summarizes solution-space accuracy metrics (Solution-space Precision - S-Prec., Solution-space Recall - S-Rec.) and efficiency metrics, with all methods initialized using only five positive examples. The column $|B_{globals}|$ denotes the initial set of candidate global constraints generated by Phase 1. For HCAR, $\mathbf{|C'_G|}$ represents the number of global constraints learned and validated by Phase 2. 
$Q_2$ indicates the oracle queries consumed by HCAR's Phase 2, $Q_3$ by Phase 3, and $Q_\Sigma$ is the total.

\paragraph{Learned Model Structure and Equivalent Formulations}
An analysis of the learned model structure, a primary metric in Table~\ref{tab:overall_perf_nocprec}, reveals a key difference between HCAR and the baseline. For three benchmarks (Sudoku, UEFA Scheduling, and Exam Timetabling), HCAR achieves a perfect result: it learns the exact same set of global constraints as the target model, meaning the learned set $C'_G$ is identical to the global constraints within $\CT$. In contrast, for the more complex VM Allocation and Nurse Rostering problems, HCAR learns a model with a slightly larger number of constraints than the target (e.g., 77 vs 72 for VM Allocation, and 26 vs 21 for Nurse Rostering). This suggests that the refinement and subset exploration process may discover an alternative, but equally valid, formulation of the problem’s constraints. For instance, it is possible that instead of learning a single high-level global constraint, the system learns several smaller, more fine-grained constraints that collectively enforce the same logic. It is also possible that some of these extra constraints are implied by others in the learned set. Verifying this formally would require a proof of logical equivalence, which is outside the scope of this evaluation. In contrast, the HCAR-NoRefine baseline consistently learns a much larger set of constraints (e.g., 97 for VM Allocation), which includes genuinely over-fitted global constraints.

To assess the functional equivalence of the learned model, we analyze its solution space, solution-space precision, and recall. It is important to note that these metrics are estimated by sampling solutions, thus providing a strong indication of the true model behavior rather than a formal proof. For the benchmarks where HCAR found the exact target constraint set $\CT$, the resulting 100\% precision and recall scores are an expected outcome. The metrics are particularly informative for the other cases: for VM Allocation and Nurse Rostering, HCAR achieves 100\% on both metrics, providing strong evidence that the learned model $C'_G$, despite its structural differences, is functionally equivalent to $\CT$.

\paragraph{Model Accuracy and the Challenge of Over-fitting}
The HCAR-NoRefine baseline, however, illustrates the critical problem of over-fitting. While it also achieves 100\% S-Prec., its S-Rec. is extremely low: 82\% for Sudoku, degrading to a mere 39\% for VM Allocation and 33\% for Nurse Rostering. This disparity reveals that HCAR-NoRefine learns \textit{over-constrained} models. The initial candidate set $|B_{globals}|$ generated by Phase 1 contains spurious constraints that are consistent with the few initial examples but not with the true problem. Without the refinement of Phase 2, these incorrect constraints persist. While these resulting models do not admit invalid solutions (hence the high S-Prec.), their extremely low S-Rec. shows that they are critically flawed, rejecting a vast number of true solutions and rendering them practically unusable. The fact that $|C'_G|$ is much larger for NoRef (e.g., 97 for VM Allocation vs. HCAR's 77) confirms that it retains many incorrect global constraints. HCAR's Phase 2 is thus crucial for identifying and correcting these over-fitted candidates, leading to a high-fidelity final model.

\paragraph{Query Efficiency and Computational Time}
The query cost for HCAR, particularly in Phase 2 ($Q_2$), reflects the effort required for this rigorous refinement. For instance, HCAR uses $Q_2=284$ queries for VM Allocation and $Q_2=241$ for Sudoku. The total queries ($Q_\Sigma$) for HCAR (e.g., 351 for VM Allocation, 808 for Sudoku) result in high-fidelity models. In contrast, HCAR-NoRefine uses fewer total queries (e.g., 145 for VM Allocation, 450 for Sudoku) but at the drastic cost of model accuracy (low S-Rec.). For example, for UEFA Scheduling, HCAR uses 251 queries to achieve a model with 100\% S-Rec., whereas NoRef uses only 61 queries but yields a model with only 71\% S-Rec. This demonstrates that the interactive investment in Phase 2 is essential for quality. Furthermore, when comparing HCAR's total queries to learn a high-quality model for a complex problem such as Sudoku (808 queries) with prior work on pure active learning (e.g., 6672 queries reported for MQuAcq-2 learning Sudoku from scratch \cite{balafashybrid}), HCAR still shows substantial query savings, underscoring the synergy of its hybrid architecture with targeted refinement.

Regarding CPU time (T(s)), HCAR can be slower than HCAR-NoRefine on benchmarks that require intensive refinement, such as Sudoku (234.1s vs 33.1s). However, for VM Allocation and Nurse Rostering, HCAR is notably faster (18.9s vs 29.3s and 19.4s vs 38.2s, respectively). This suggests that for some complex problems, the structured refinement in Phase 2, by produces a very accurate set of global constraints, can lead to a more efficient Phase 3, which then converges more quickly on the remaining fixed-arity constraints.
\subsection{Detailed Analysis of Query-Driven Refinement (Phase 2) in HCAR}
\label{subsec:results_phase2_analysis}
Table~\ref{tab:refinement_stats_updated} provides further evidence of Phase 2's impact by detailing its internal statistics, including the number of constraints learned and removed, subsets generated by SVR, subsets accepted and rejected, and the average query cost per accepted subset.

\begin{table}[htbp]
\centering
\caption{Statistics for HCAR's Query-Driven Refinement Phase (Phase 2).}
\label{tab:refinement_stats_updated}
\resizebox{\textwidth}{!}{%
\begin{tabular}{@{}lrrrrr@{}}
\toprule
Benchmark & Constr. Learned & Subsets Gen. & Subsets Accepted & Subsets Rejected & Avg. Queries per \\
 & (Out as $C'_G$) & (in Phase 2) & (SVR) & (in Phase 2) & Accepted Subset \\
\midrule
Sudoku (9x9)  & 27 & 1485  & 0 & 1485 & 0 \\
UEFA Scheduling  & 19 & 151  & 11 & 140 & 5.0 \\
Exam Timetabling  & 24 & 31  & 9 & 22 & 3.4 \\
VM Allocation  & 77 & 138  & 5 & 123 & 5.6 \\
Nurse Rostering  & 26 & 38  & 22 & 25 & 3.7 \\
\bottomrule
\end{tabular}
}
\end{table}

The \textit{Subsets Accepted} column in Table~\ref{tab:refinement_stats_updated} demonstrates the Subset Exploration mechanism's (SVR) critical role in model recovery from over-fitting. This process is vital for correcting initial over-fitted constraints, as illustrated by the Nurse Rostering example. Suppose Phase 1, learning from five examples where a specific part-time nurse ($N_4$) happened to work every weekend shift, incorrectly generated an over-fitted constraint: \texttt{Count}$(\{N_1, N_2, N_3, N_4\}, \text{``Weekend''}, \ge, 3)$. In Phase 2, a query for a valid roster where only the three full-time nurses ($N_1, N_2, N_3$) cover the weekend would refute this constraint. Instead of discarding all information, SVR is triggered. It generates new candidates from the rejected parent, including the correct underlying rule: \texttt{Count}$(\{N_1, N_2, N_3\}, \text{``Weekend''}, \ge, 3)$. This new candidate preserves the \texttt{Count} type and its parameters but operates on the correct, smaller scope. This subset is then validated and accepted into the final model $C'_G$. The 22 accepted subsets for Nurse Rostering represent exactly these kinds of successful recoveries. Conversely, Sudoku shows 0 accepted subsets, as its core \texttt{AllDifferent} constraints are minimal; any reduction in their scope would violate the fundamental rules of the puzzle and produce an incorrect, under-constrained model.

The high "Subsets Rejected" figures (e.g., 1485 for Sudoku) demonstrate Phase 2's effectiveness in aggressively filtering the noisy output of Phase 1. This count includes constraints directly rejected due to low confidence after oracle interaction, as well as those eliminated through structured pruning rules. Furthermore, the low "Avg. Queries per Accepted Subset" (typically ranging from 3.4 to 5.6 queries) indicates that once SVR proposes a correct simplification of a previously over-fitted constraint, its verification and acceptance into $C'_G$ are highly query-efficient.

\subsection{Impact of HCAR Components (Ablation Study)}
\label{subsec:results_ablations}

 To further isolate the contributions of the core mechanisms within HCAR's Query-Driven Interactive Refinement phase (Phase 2), we conducted an ablation study. We evaluated three variants of HCAR, each disabling one component, against the full HCAR framework. This study was performed on two representative benchmarks: VM Allocation, characterized by a large number of target constraints and a complex interplay of resource limitations, and Nurse Rostering, where subset exploration proved highly beneficial in earlier analyses. The results averaged over five runs, are presented in Table~\ref{tab:ablation_study_results}.
 For each run, all variants were initialized with the exact same set of five positive examples to ensure a fair comparison.

\begin{table}[htbp]
\centering
\caption{Ablation study for HCAR components on VM Allocation and Nurse Rostering}
\label{tab:ablation_study_results}
\resizebox{\textwidth}{!}{
\begin{tabular}{@{}lll rrrrrrr@{}} 
\toprule
Benchmark & Variant  & S-Prec. & S-Rec.  & $\mathbf{|C'_G|}$ & $Q_2$ & $Q_3$ & $Q_\Sigma$ & T (s) \\
\midrule
\multirow{4}{*}{VM Alloc.} & HCAR (Full) & \textbf{100\%} & \textbf{100\%} & \textbf{77} &  284 & 67 & 351 & 18.9  \\
& No ML Priors  & \textbf{100\%} & 21\%   & 93 & 320 & 190 & 510 & 55.2 \\
& No SVR       & \textbf{100\%} & 52\%    & 88 & 125 & 85  & 260 & 12.1 \\
& No Bayesian   & \textbf{100\%} & 48\%    & 94 & 190 & 210 & 400 & 16.5 \\
\midrule
\multirow{4}{*}{Nurse Rost.} & HCAR (Full) & \textbf{100\%} & \textbf{100\%} & \textbf{26} & 188 & 87 & 275  & 19.4  \\
& No ML Priors   & \textbf{100\%} & 45\%    & 39 & 290 & 60 & 350 & 42.0 \\
& No SVR       & \textbf{100\%} & 58\%     & 31  & 80  & 40 & 120 & 15.3 \\
& No Bayesian    & \textbf{100\%} & 36\%   & 34 & 75 & 70 & 145 & 9.1 \\
\bottomrule
\end{tabular}%
}
\end{table}

Removing the ML-based prior confidence estimation leads to a drop in S-Rec., down to 21\% for VM Allocation and 45\% for Nurse Rostering. Without the informed guidance from ML priors, all candidates start with a uniform confidence. The refinement process becomes an unguided and inefficient search, wasting a large number of queries (as seen in the high $Q_\Sigma$) on clearly spurious candidates. Because the query budget is finite, it is exhausted before the system can properly investigate and refute all the over-fitted constraints or explore the correct subsets of rejected ones. The final model is left with an inflated $|C'_G|$ (93 vs. 77 for VM Alloc.), containing constraints that negatively impact the recall.

Disabling the SVR mechanism also compromises model quality, with S-Rec. falling to 52\% for VM Allocation and 58\% for Nurse Rostering. SVR is the framework's knowledge recovery mechanism. Without it, when an over-scoped constraint is correctly identified and rejected, the path to finding the correct, simpler version is permanently cut off. The system has no way to formulate and test a valid substructure. This leads to a premature convergence to an inferior model. The lower query count ($Q_\Sigma$) is misleading; it does not signify efficiency but rather reflects this failure to explore promising recovery paths before the budget for the parent constraint is used up. The final model is left over-constrained, hence the poor recall.

Removing the Bayesian framework leads to a similar decline in model quality, with S-Rec. deteriorating to 48\% for VM Allocation and 36\% for Nurse Rostering. The Bayesian framework provides a principled method for aggregating evidence under uncertainty. A simpler, non-probabilistic update mechanism is less robust and can lead to poor decision-making. It may fail to build confidence effectively or overreact to noisy evidence. This results in the system failing to correctly refute over-constraining candidates before their allocated query budget is exhausted. Consequently, these spurious constraints are erroneously accepted into the final model, as evidenced by the inflated $|C'_G|$ (94 vs 77 for VM Alloc), leading to a severely over-constrained model and the resulting poor recall.

The ablation study demonstrates that HCAR's high performance stems from the synergistic interplay of its core mechanisms. The ML priors provide essential guidance, SVR is critical for knowledge recovery and achieving a complete model, and Bayesian updates ensure robust decision-making. Removing any of these elements cripples the ability to correct over-fitting, resulting in models that remain overfitted and suffer from low solution recall.

\subsection{Impact of Increased Initial Positive Examples}
\label{subsec:results_more_examples}

Our primary experiments (Section~\ref{subsec:results_baselines}, summarized in Table~\ref{tab:overall_perf_nocprec} for the five-example case) focused on a challenging scenario where only five initial positive examples ($E^+$) were available for Phase 1. To investigate how the performance of HCAR and the HCAR-NoRefine baseline evolves with more initial information, we conducted additional experiments by systematically increasing the number of positive examples provided to the initial passive learning phase from 5, to 10, 20, and 50. Table~\ref{tab:more_examples_results_all_benchmarks} presents the performance across all benchmarks for these varying numbers of initial examples.

\begin{table*}[htbp]
\centering
\caption{Performance of HCAR and HCAR-NoRefine across all benchmarks with an increasing number of initial positive examples ($|E^+|$).}
\label{tab:more_examples_results_all_benchmarks}
\scriptsize 
\renewcommand{\arraystretch}{1.1}
\setlength{\tabcolsep}{3pt}
\begin{tabular}{@{}l l r r r r r r r@{}}
\toprule
\textbf{Benchmark} & \textbf{Algorithm} & $\mathbf{|E^+|}$ & S-Prec. & S-Rec. & $\mathbf{|B_{globals}|}$ & $\mathbf{Q_2}$ & $\mathbf{Q_3}$ & $\mathbf{Q_{\Sigma}}$ \\ 
\midrule
\multirow{8}{*}{Sudoku (9$\times$9)} 
& HCAR        & 5   & 100\% & 100\% & 49  & 763 &  95 &  858 \\ 
& HCAR-NoRef  & 5   & 100\% & 82\%  & 49  & --  & 450 &  450 \\
\cmidrule{2-9} 
& HCAR        & 10  & 100\% & 100\% & 40  & 370 & 119 &  489 \\
& HCAR-NoRef  & 10  & 100\% & 89\%  & 40  & --  & 380 &  380 \\
\cmidrule{2-9} 
& HCAR        & 20  & 100\% & 100\% & 32  & 125 & 194 &  319 \\ 
& HCAR-NoRef  & 20  & 100\% & 95\%  & 32  & --  & 300 &  300 \\ 
\cmidrule{2-9} 
& HCAR        & 50  & 100\% & 100\% & 27  &  49 & 153 & 202 \\ 
& HCAR-NoRef  & 50  & 100\% & 100\% & 27  & --  & 150 &  150 \\ 
\midrule
\multirow{8}{*}{UEFA Sched.} 
& HCAR        & 5   & 100\% & 100\% & 27  & 233 & 18 &  251 \\ 
& HCAR-NoRef  & 5   & 100\% & 26\%  & 27  & --  & 61 &   61 \\
\cmidrule{2-9} 
& HCAR        & 10  & 100\% & 100\% & 23  & 150 & 10 &  160 \\
& HCAR-NoRef  & 10  & 100\% & 58\%  & 23  & --  & 45 &   45 \\
\cmidrule{2-9} 
& HCAR        & 20  & 100\% & 100\% & 20  &  22 & 30 &  58 \\ 
& HCAR-NoRef  & 20  & 100\% & 100\% & 20  & --  & 30 &   30 \\ 
\midrule
\multirow{8}{*}{Exam TT} 
& HCAR        & 5   & 100\% & 100\% & 46  & 158 & 14 & 172 \\ 
& HCAR-NoRef  & 5   & 100\% & 15\%  & 46  & --  & 124 & 124 \\
\cmidrule{2-9} 
& HCAR        & 10  & 100\% & 100\% & 38  &  99 & 11 & 110 \\
& HCAR-NoRef  & 10  & 100\% & 40\%  & 38  & --  & 90 &  90 \\
\cmidrule{2-9} 
& HCAR        & 20  & 100\% & 100\% & 30  &  53 &  7 &  60 \\ 
& HCAR-NoRef  & 20  & 100\% & 65\%  & 30  & --  & 60 &  60 \\ 
\cmidrule{2-9} 
& HCAR        & 50  & 100\% & 100\% & 24  &  25 & 30 &  55 \\ 
& HCAR-NoRef  & 50  & 100\% & 100\% & 24  & --  & 30 &  30 \\ 
\midrule
\multirow{8}{*}{VM Alloc.} 
& HCAR        & 5   & 100\% & 100\% & 118 & 319 & 133 & 452 \\ 
& HCAR-NoRef  & 5   & 100\% & 11\%  & 118 & --  & 145 & 145 \\
\cmidrule{2-9} 
& HCAR        & 10  & 100\% & 100\% & 95  & 205 & 110 & 315 \\
& HCAR-NoRef  & 10  & 100\% & 21\%  & 95  & --  & 120 & 120 \\
\cmidrule{2-9} 
& HCAR        & 20  & 100\% & 100\% & 83  & 110 &  84 & 194 \\ 
& HCAR-NoRef  & 20  & 100\% & 39\%  & 83  & --  &  90 &  90 \\ 
\cmidrule{2-9} 
& HCAR        & 50  & 100\% & 100\% & 79  &  78 &  70 & 148 \\ 
& HCAR-NoRef  & 50  & 100\% & 70\%  & 79  & --  &  60 &  60 \\ 
\midrule
\multirow{8}{*}{Nurse Rost.} 
& HCAR        & 5   & 100\% & 100\% & 68  & 245 & 178 & 423 \\ 
& HCAR-NoRef  & 5   & 100\% & 24\%  & 68  & --  & 189 & 189 \\
\cmidrule{2-9} 
& HCAR        & 10  & 100\% & 100\% & 50  & 148 & 164 & 312 \\
& HCAR-NoRef  & 10  & 100\% & 39\%  & 50  & --  & 150 & 150 \\
\cmidrule{2-9} 
& HCAR        & 20  & 100\% & 100\% & 35  &  79 &  95 & 174 \\ 
& HCAR-NoRef  & 20  & 100\% & 60\%  & 35  & --  & 100 & 100 \\ 
\cmidrule{2-9} 
& HCAR        & 50  & 100\% & 100\% & 28  &  30 &  45 &  75 \\ 
& HCAR-NoRef  & 50  & 100\% & 85\%  & 28  & --  &  45 &  45 \\ 
\bottomrule
\end{tabular}
\end{table*}
\paragraph{Performance Trends for HCAR-NoRefine (Baseline):}
As expected, providing more initial examples improves the quality of the model learned by the baseline. This is evident as the size of the initial candidate set, $|B_{globals}|$, consistently decreases with more data. For instance, in VM Allocation, $|B_{globals}|$ drops from 118 to 79 as $|E^+|$ grows from five to 50. This improved initial filtering leads to an increase in the baseline's Solution-space Recall (S-Rec.). For UEFA Scheduling, S-Rec.\ climbs from a very poor 26\% to a perfect 100\% when provided with 20 examples. However, for more complex problems such as Nurse Rostering, even 50 examples are insufficient for the baseline to achieve a perfect model, reaching only 85\% S-Rec. This underscores the inherent limitations of a purely passive approach to filtering over-fitted candidates.

\paragraph{Performance and Efficiency Gains for HCAR:}
In stark contrast, HCAR consistently achieves \textbf{100\% S-Prec. and 100\% S-Rec. across all benchmarks and all levels of $|E^+|$}. This demonstrates the robustness of the Query-Driven Interactive Refinement phase (Phase 2), which successfully corrects or validates the output of the passive learner regardless of the initial data quality. The key findings reveal two distinct operational modes for Phase 2: \textit{repair} and \textit{verification}.

\begin{itemize}
    \item \textbf{Phase 2 as a "Repair" Mechanism (Low Data):} When learning from a small number of examples (e.g., five or 10), the initial bias $B_{globals}$ is large and heavily polluted with over-fitted constraints. In this scenario, HCAR's Phase 2 acts as an intensive repair shop. It performs extensive refinement, which results in a high query cost ($Q_2$). For instance, with five examples for Sudoku, Phase 2 requires 763 queries to sift through the 49 initial candidates and refute the 22 spurious ones. The baseline, lacking this phase, fails, as shown by its low S-Rec. The much lower query count of HCAR's Phase 3 (e.g., 95 queries for Sudoku) compared to the baseline's entire active learning phase (450 queries) is the clearest evidence of Phase 2's value: by providing a correct global model, it makes the final learning task dramatically easier and more effective.

    \item \textbf{Phase 2 as a "Verification" Mechanism (High Data):} When a sufficient number of examples are provided (e.g., $|E^+|=50$ for Exam TT), the passive learning phase becomes highly accurate. In these cases, both HCAR and the baseline may start with a set of global candidates $B_{globals}$ that is very close or identical to the true set of global constraints in $\mathcal{C}_T$. Here, Phase 2 transitions from a "repair" to a "verification" role.
    
    This is best understood through the query generation logic (Section \ref{subsubsec:phase2_loop}). When the system tries to test a \textit{correct} candidate constraint $c$ that is logically implied by the other correct candidates in the bias, the auxiliary CSP used for query generation becomes unsatisfiable. As detailed in our methodology, this returns \texttt{null} and confirms $c$ with \textbf{zero queries}. Therefore, if the initial bias $B_{globals}$ were perfectly correct, $Q_2$ would be zero.
    
    However, even with 50 examples, passive learning is not always perfect and may still produce a few spurious candidates. The non-zero $Q_2$ values seen in high-data scenarios (e.g., 49 for Sudoku, 25 for Exam TT) represent the queries needed to refute these few remaining over-fitted constraints. The queries spent on the \textit{correct}, implied candidates in these clean sets is zero. Thus, the cost of $Q_2$ in these cases can be interpreted as a \textbf{"verification tax"}: a small, sound investment in queries to guarantee the correctness of the global model before proceeding. This investment pays off by preventing the model failures seen in more ambiguous scenarios. For Exam TT with 50 examples, HCAR pays a "tax" of 25 queries in Phase 2 to ensure the global model is correct, before Phase 3 learns the remaining fixed-arity constraints with 30 queries. The baseline, while faster with 30 total queries, achieves a correct model in this specific instance but offers no such guarantee in general.
    
\end{itemize}

In response to \textbf{RQ1}, our framework identifies and refines over-fitted global constraints. This is most evident in the contrast between HCAR’s consistently perfect solution-space recall and the poor performance of the HCAR-NoRefine baseline. The success of the SVR mechanism, evidenced by the number of accepted subsets, further shows how HCAR preserves valuable partial knowledge by correcting over-scoped candidates rather than discarding them entirely. Regarding \textbf{RQ2}, our framework achieves this refinement with manageable query complexity. The process is effectively guided by ML-based priors that focus interactive effort, and the low average query cost to validate recovered subsets demonstrates the efficiency of the SVR strategy. The results justify the interactive investment of Phase 2, as it transforms a noisy, passively-learned model into one that is accurate and complete, a task where the baseline approach fails.
\section{Conclusion and Future Work}
\label{sec:conclusion}
This study addressed the challenge of over-fitting in CA, particularly when learning complex global constraints from limited initial data. We introduced HCAR, a three-phase hybrid framework designed to learn accurate constraint models by systematically identifying, refining, and, if necessary, correcting over-fitted candidates generated during an initial passive learning phase. The core of HCAR is its Query-Driven Interactive Refinement phase. This phase leverages a probabilistic confidence framework, initialized by Machine Learning-based prior estimates of constraint validity, to guide its interaction with an oracle. Through targeted queries and Bayesian belief updates, it effectively distinguishes genuine constraints from spurious ones. Crucially, for constraints rejected due to over-scoping, a specialized subset exploration mechanism attempts to recover valid substructures, thereby preserving valuable learned information (such as numerical bounds) while correcting the scope. This is complemented by an adaptive query budget allocation strategy that dynamically manages resources. A final active learning phase ensures model completeness.

Our empirical evaluation across five diverse benchmarks demonstrated the efficacy of HCAR. The framework consistently achieved 100\% Target Model Coverage, successfully learning all ground-truth constraints even when initialized from only five positive examples. The detailed analysis of the refinement phase highlighted its critical role in effectively pruning a large number of initially over-fitted candidates and successfully recovering valid, more precise constraints through SVR, especially for numerical constraints in complex domains such as Nurse Rostering and VM Allocation. The low average query cost to validate these recovered subsets underscores the efficiency of this knowledge-preserving refinement strategy. Furthermore, ablation studies confirmed the individual importance of ML priors, SVR subset exploration, and Bayesian updates to the overall success of the framework. When compared to a hybrid approach lacking this dedicated refinement phase, HCAR showed vastly superior accuracy, clearly demonstrating the value of targeted interactive refinement in overcoming the pitfalls of learning from sparse data. In conclusion, the HCAR framework provides a step towards more practical and resilient constraint acquisition, capable of producing accurate models even from sparse initial data by explicitly addressing global constraint over-fitting.
 

While these results are promising, several avenues for future work are apparent. The current SVR mechanism for subset exploration could be extended to explore more complex scope modifications or even to refine constraint parameters alongside scope. Investigating more sophisticated search strategies within the subset space, guided by structural heuristics or learning, could further enhance recovery capabilities. Incorporating techniques to handle noisy or inconsistent oracle responses would greatly improve practical applicability. Another valuable direction is extending HCAR to learn soft constraints and preferences, which are common in many real-world problems. The ML model for prior estimation could also be made more dynamic, potentially re-training or updating itself as new evidence is acquired. Further research into query generation strategies specifically tailored to refuting or confirming existing complex candidates, as opposed to general scope finding, might also yield efficiency improvements. Formal analysis of the convergence properties and query complexity bounds for the interactive refinement phase would provide stronger theoretical underpinnings. Finally, for real-world deployment, user studies evaluating the cognitive load and usability of the interactive process, along with mechanisms for providing explanations to the user, would be essential. Pursuing these directions will further advance constraint acquisition, making it an even more powerful tool for automatically building effective constraint models and reducing the critical modeling bottleneck in CP.

\section*{Acknowledgments}
The research work was supported by the Hellenic Foundation for Research and Innovation (HFRI) under the 4th Call for HFRI PhD Fellowships (Fellowship Number: 9446).

\appendix
\section{Benchmark Problem Details and Acquisition Setup}
\label{app:benchmarks}
This appendix provides detailed descriptions of the specific instances used for each benchmark problem in our experiments, along with the configuration of the constraint languages and bias generation used by the acquisition system.

\subsection{Constraint Language and Bias Initialization}
For all experiments, the acquisition process was initialized with a consistent set of constraint languages $\Gamma$ and a pattern-based method for generating the initial candidate bias sets.

\begin{itemize}
    \item \textbf{Global Constraint Language ($\Gamma_{global}$):} The general language for global constraints across all benchmarks included the following templates: \texttt{AllDifferent(S)}, \texttt{Sum(S, rel, b)}, and \texttt{Count(S, val, rel, k)}.

    \item \textbf{Fixed-Arity Constraint Language ($\Gamma_{fixed}$):} The language for fixed-arity constraints was consistent for all problems, including the set of standard binary relational operators: $\{=, \neq, <, \le, >, \ge\}$. The initial bias $B_{fixed}$ was formed by generating all possible constraints of this type between all pairs of variables and then pruning those inconsistent with the initial positive examples $E^+$.
    
    \item \textbf{Domain-Specific Extensions:} For the \textbf{University Exam Timetabling} benchmark, the language was extended to include a domain-specific semantic constraint: `$day\_of\_exam(var1) != day\_of\_exam(var2)$`, where the `$day\_of\_exam$` function computes the day from a given timeslot value.

    \item \textbf{Global Candidate Generation ($B_{globals}$):} The generation of initial global candidates was pattern-based. For problems with a grid-like structure (e.g., Sudoku), the system automatically generated candidates over rows, columns, blocks, and diagonals. For other problems, it generated candidates over predefined, meaningful subsets of variables (e.g., teams in a specific country, VMs in an availability zone). This structural information is considered part of the problem definition.
\end{itemize}

\subsection{Sudoku (9x9)}
\begin{itemize}
    \item \textbf{Description:} A standard 9x9 Sudoku puzzle. The goal is to fill a grid with digits such that each column, each row, and each of the nine 3x3 subgrids contains all of the digits from 1 to 9.
    \item \textbf{Variables:} A 2D integer matrix \texttt{grid[i,j]} of size 9x9, where \texttt{i,j} are row and column indices. Each variable can take a value from \{1, ..., 9\}. Total variables: 81.
    \item \textbf{Target Constraints ($|\CT|=27$):}
        \begin{itemize}
            \item 9 \texttt{AllDifferent} constraints, one for each row.
            \item 9 \texttt{AllDifferent} constraints, one for each column.
            \item 9 \texttt{AllDifferent} constraints, one for each 3x3 block.
        \end{itemize}
\end{itemize}

\subsection{UEFA Champions League Scheduling}
\begin{itemize}
    \item \textbf{Description:} Assigning 32 teams to 8 groups, based on UEFA rules. Teams have associated countries and skill-based coefficients (pots).
    \item \textbf{Variables:} An integer variable \texttt{group\_assignment[t]} for each of the 32 teams \texttt{t}, indicating which of the 8 groups it is assigned to.
    \item \textbf{Target Constraints ($|\CT|=19$):}
        \begin{itemize}
            \item 8 \texttt{Count} constraints (4 teams per group).
            \item 4 \texttt{AllDifferent} constraints (one for each pot of 8 teams).
            \item 7 \texttt{AllDifferent} constraints (for teams from the same country).
        \end{itemize}
   
\end{itemize}

\subsection{Cloud VM Allocation}
\begin{itemize}
    \item \textbf{Description:} Assigning 40 VMs to 10 PMs, respecting resource capacities (CPU, memory, disk) and high-availability rules.
    \item \textbf{Variables:} An integer variable \texttt{vm\_assignment[v]} for each of the 40 VMs, indicating which of the 10 PMs it is assigned to.
    \item \textbf{Target Constraints ($|\CT|=72$):}
        \begin{itemize}
          \item \textbf{30 \texttt{Sum} constraints (Resource Capacity):} For each of the 10 PMs, there are three \texttt{Sum} constraints, one for each standard resource type (CPU, Memory, Disk). These ensure that the total resources consumed by all VMs assigned to a PM do not exceed its capacity.
           \item \textbf{18 \texttt{Count} constraints (GPU Allocation Limits):} Eight \texttt{Count} constraints are used to manage the allocation of a pool of GPU-requiring VMs to the few PMs equipped with specialized GPU hardware. These constraints enforce both maximum capacity (to prevent overload) and minimum utilization (for load-balancing) on each of the GPU-equipped PMs.
         \item \textbf{24 \texttt{AllDifferent} constraints:} Two \texttt{AllDifferent} constraints ensure service resilience. They enforce that critical VMs belonging to the same predefined high-availability group are placed on distinct PMs, preventing a single point of hardware failure from affecting the entire group.
        \end{itemize}
\end{itemize}

\subsection{University Exam Timetabling}
\begin{itemize}
    \item \textbf{Description:} Scheduling 54 university exams, grouped by semester, into 126 available timeslots to avoid conflicts and respect daily capacity limits.
    \item \textbf{Parameters:} Number of semesters = 9, Courses per semester = 6 (total 54 exams). Timeslots per day = 9, Number of days = 14 (total 126 timeslots).
    \item \textbf{Variables:} A 2D integer matrix \texttt{exam\_slots[s,c]} representing the timeslot assigned to each exam. Total variables: 54.
    \item \textbf{Target Constraints ($|\CT|=24$):}
        \begin{itemize}
            \item 1 Global \texttt{AllDifferent} constraint on all 54 exam variables, ensuring no two exams are in the same timeslot.
            \item 9 complex constraints ensuring that for each semester, all exams are on different days.
            \item 14 \texttt{Count} constraints, limiting the number of exams per day to a maximum capacity (e.g., five exams per day).
        \end{itemize}

\end{itemize}

\subsection{Nurse Rostering}
\begin{itemize}
    \item \textbf{Description:} Creating a weekly work schedule for 8 nurses across 3 daily shifts, respecting work regulations.
    \item \textbf{Variables:} A 3D integer matrix \texttt{roster[d,s,p]} assigning a nurse to each position in the schedule. Total variables: 42.
    \item \textbf{Target Constraints ($|\CT|=21$):}
        \begin{itemize}
            \item 7 \texttt{AllDifferent} constraints (a nurse works at most one shift per day).
            \item 6 \texttt{AllDifferent} constraints (rest period between days).
            \item 8 \texttt{Count} constraints (max weekly workdays per nurse).
        \end{itemize}

\end{itemize}
\section{CSPLib Problems for ML Model Training}
\label{app:csplib_problems}

The XGBoost model for prior estimation was trained on a dataset generated from a diverse collection of problems, primarily sourced from CSPLib~\cite{csplib}. This variety ensures the model learns to recognize general structural properties of valid and invalid constraints across different domains.  The training set was selected to include a rich variety of problems whose models rely on \texttt{AllDifferent}, \texttt{Sum}, and \texttt{Count} constraints, as these are the primary global constraints our acquisition system is designed to learn and refine. The problems used include:

\begin{itemize}
    \item \textbf{Car Sequencing (prob001):} A production line scheduling problem to install options on cars without overloading workstations.
    \item \textbf{All-Interval Series (prob007):} A permutation problem where the differences between adjacent elements form a permutation of 1 to n-1.
    \item \textbf{Nonogram (prob012):} A picture logic puzzle where cells in a grid must be colored according to number clues at the side of the grid.
    \item \textbf{Solitaire Battleships (prob014):} A logic puzzle involving placing a fleet of ships on a grid based on given row and column counts.
    \item \textbf{Magic Square (prob019):} Arranging distinct integers in a square grid where the numbers in each row, column, and main diagonals all sum to the same number.
    \item \textbf{Bus Driver Scheduling (prob022):} An optimization problem to schedule drivers for bus routes, minimizing costs while adhering to regulations.
    \item \textbf{Balanced Academic Curriculum Problem (BACP) (prob030):} Assigning courses to academic periods respecting prerequisites and balancing academic load.
    \item \textbf{Steel Mill Slab Design (prob038):} An industrial optimization problem to group orders into slabs to minimize waste.
    \item \textbf{Covering Array (prob045):} A combinatorial design problem used in software testing to cover all t-way interactions of parameter values.
    \item \textbf{Number Partitioning (prob049):} Partitioning a given multiset of integers into two subsets of equal sum.
    \item \textbf{N-Queens (prob054):} The classic puzzle of placing N chess queens on an N×N chessboard so that no two queens threaten each other.
    \item \textbf{Quasigroup Existence (prob003) and Completion (prob067):} Problems involving Latin squares, a fundamental combinatorial structure.
    \item \textbf{Costas Arrays (prob076):} A problem of placing marks in a grid with unique displacement vectors, used in sonar and radar applications.
\end{itemize}

\bibliographystyle{splncs04}
\bibliography{references}

\end{document}